\documentclass[10pt,twocolumn,letterpaper]{article}

\usepackage{cvpr}              

\usepackage{times}
\usepackage{epsfig}
\usepackage{graphicx}
\usepackage{amsmath}
\usepackage{amssymb}

\usepackage[caption = false]{subfig}
\usepackage{comment}
\usepackage{color}

\usepackage{dsfont}
\usepackage{multirow}
\usepackage{cite}

\usepackage{array}
\usepackage{booktabs, caption, makecell}

\usepackage{algorithm}
\usepackage{algpseudocode}

\usepackage[pagebackref=true,breaklinks=true,letterpaper=true,colorlinks,bookmarks=false]{hyperref}

\newcolumntype{L}[1]{>{\raggedright\arraybackslash}m{#1}}
\newcolumntype{C}[1]{>{\centering\arraybackslash}m{#1}}
\newcolumntype{R}[1]{>{\raggedleft\arraybackslash}m{#1}}
\newcolumntype{+}{>{\global\let\currentrowstyle\relax}}
\newcolumntype{^}{>{\currentrowstyle}}

\newcommand{\RomNum}[1]{\MakeUppercase{\romannumeral #1}}

\newcommand{\bfl}{\ensuremath{{\mathbf{l}}}}

\newcommand{\bfx}{\ensuremath{{\mathbf{x}}}}

\newcommand{\bfc}{\ensuremath{{\mathbf{c}}}}
\newcommand{\bfu}{\ensuremath{{\mathbf{u}}}}
\newcommand{\bfv}{\ensuremath{{\mathbf{v}}}}
\newcommand{\bfU}{\ensuremath{{\mathbf{U}}}}
\newcommand{\bfA}{\ensuremath{{\mathbf{A}}}}
\newcommand{\bfV}{\ensuremath{{\mathbf{V}}}}

\usepackage[capitalize]{cleveref}
\crefname{section}{Sec.}{Secs.}
\Crefname{section}{Section}{Sections}
\Crefname{table}{Table}{Tables}
\crefname{table}{Tab.}{Tabs.}

\def\cvprPaperID{6918} 
\def\confName{CVPR}
\def\confYear{2022}

\begin{document}

\title{Eigenlanes: Data-Driven Lane Descriptors for Structurally Diverse Lanes}

\author{Dongkwon Jin$^1$, Wonhui Park$^1$, Seong-Gyun Jeong$^2$, Heeyeon Kwon$^2$, Chang-Su Kim$^1$\\
Korea University$^1$, 42dot.ai$^2$\\
{\tt\footnotesize \{dongkwonjin, whpark\}@mcl.korea.ac.kr, \tt\footnotesize \{seonggyun.jeong, heeyeon.kwon\}@42dot.ai,
\tt\footnotesize changsukim@korea.ac.kr}
}
%

\maketitle

\begin{abstract}
    A novel algorithm to detect road lanes in the eigenlane space is proposed in this paper. First, we introduce the notion of eigenlanes, which are data-driven descriptors for structurally diverse lanes, including curved, as well as straight, lanes. To obtain eigenlanes, we perform the best rank-$M$ approximation of a lane matrix containing all lanes in a training set. Second, we generate a set of lane candidates by clustering the training lanes in the eigenlane space. Third, using the lane candidates, we determine an optimal set of lanes by developing an anchor-based detection network, called SIIC-Net. Experimental results demonstrate that the proposed algorithm provides excellent detection performance for structurally diverse lanes. Our codes are available at \href{https://github.com/dongkwonjin/Eigenlanes}{https://github.com/dongkwonjin/Eigenlanes}.
\end{abstract}

\section{Introduction}

\begin{figure}[t]
  \centering
  \includegraphics[width=1\linewidth]{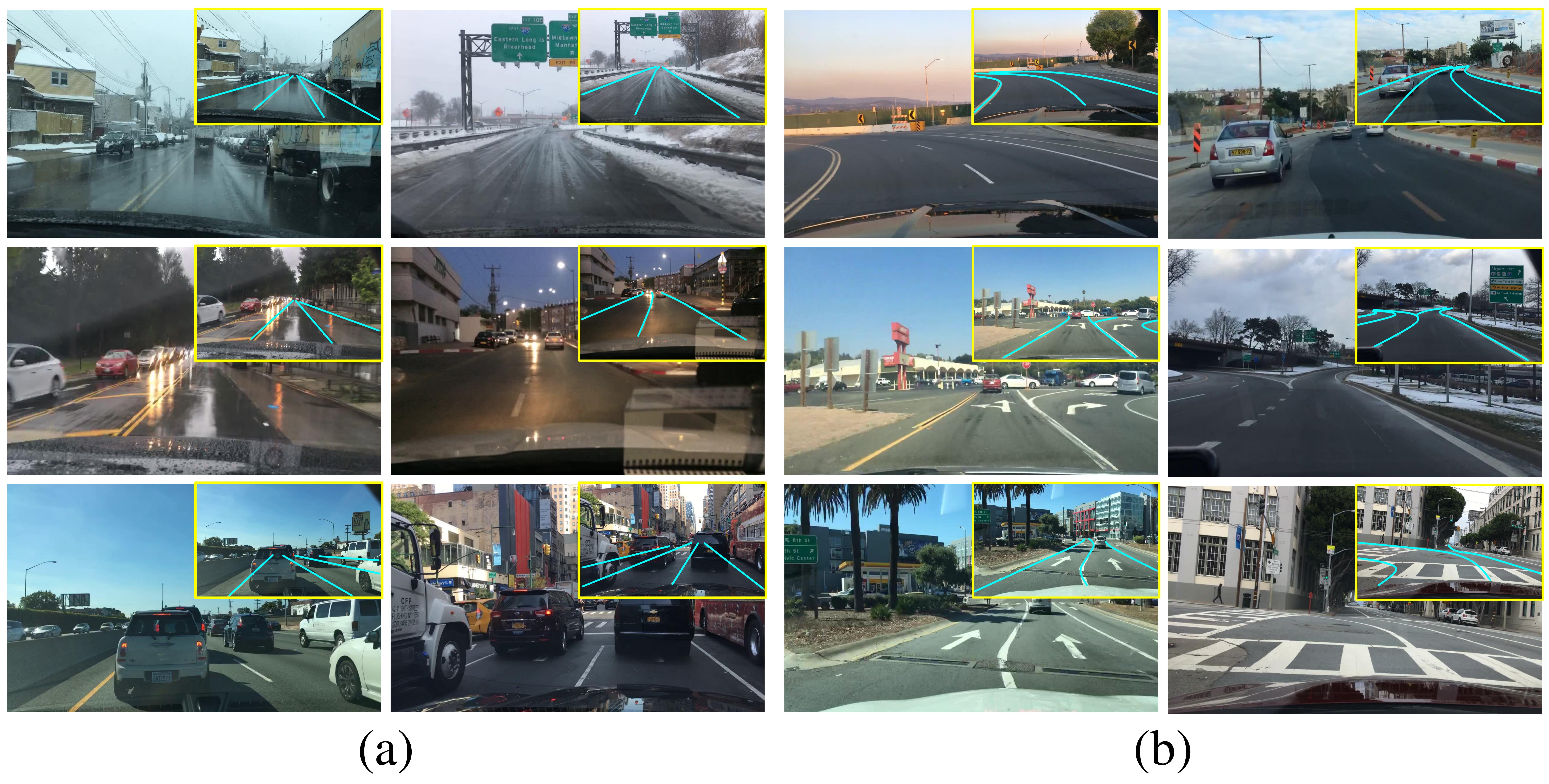}
  \setlength{\abovecaptionskip}{-0.3cm}
  \caption{There are two challenging factors in lane detection. First, in (a), lanes may be implicit due to adverse weather conditions or occlusion by nearby vehicles. Second, in (b), it is difficult to design lane anchors due to the structural diversity of lanes. The ground-truth lanes are shown in cyan within the insets.}
  \label{fig:Challenge}
\end{figure}

Lane detection is essential for understanding driving environments, in which autonomous or human drivers should abide by traffic rules. To control vehicle maneuvers, boundaries of road lanes and sidewalks should be detected reliably. There are various challenging factors to interfere with the detection of those lanes. For example, lanes may be unobvious or even invisible due to weather and illumination conditions or due to the occlusion by nearby vehicles, as illustrated in Figure \ref{fig:Challenge}(a).

For lane detection, traditional methods extract hand-crafted features, such as image gradients or color features \cite{he2004,aly2008,hillel2014,zhou2010}. This feature engineering may fail to discriminate actual lanes from noisy ones. Recently, various CNN-based techniques have been developed to detect lanes in real environments more reliably. Most such techniques adopt the semantic segmentation framework \cite{pan2018,neven2018,ghafoorian2018gan,hou2019_road,hou2020_inter,zheng2021}, in which each pixel in an image is dichotomized into either lane or no-lane category. To preserve continuous lane structure in detection results, several attempts have been made, including curve fitting \cite{neven2018}, polynomial regression \cite{wang2020poly}, and adversarial training \cite{ghafoorian2018gan}. However, even these algorithms may fail to detect less visible lanes in cluttered scenes, because they use only local features and may miss parts of lanes. Meanwhile, the anchor-based detection framework, which has been used extensively in object detection \cite{redmon2016yolo,liu2016ssd,ren2015faster,law2018cornernet}, has been recently adopted for lane detection \cite{li2019line,tabelini2021}. These anchor-based algorithms consider straight lines as anchors (or lane candidates). Then, they declare each anchor as a lane or not. By exploiting long-range contextual information, they can detect implied lanes effectively. However, because of the straight lane assumption, it may not deal with complicated lanes, such as curved and winding ones in Figure \ref{fig:Challenge}(b).

In this paper, we propose a novel algorithm to detect structurally diverse road lanes in the eigenlane space. It enables to process curved, as well as straight, lanes reliably in the anchor-based detection framework.
First, we introduce the notion of eigenlanes, which are data-driven lane descriptors. To obtain eigenlanes, we approximate a lane matrix, which contains all lanes in a training set, based on the low-rank approximation property of singular value decomposition (SVD) \cite{y2015_SVD_Hopcroft}. Then, each lane is represented by a linear combination of $M$ eigenlanes. Second, we generate a set of lane candidates, including complicated and curved ones, by clustering the training lanes in the eigenlane space. Third, we develop an anchor-based detector, called SIIC-Net, to detect lanes from the candidates. It consists of two modules: self-lane identification (SI) module and inter-lane correlation (IC) module. SI computes the classification probability and regression offset of each candidate, while IC estimates the compatibility between each pair of lanes. Extensive experiments show that the proposed algorithm provides competitive results on existing datasets \cite{pan2018, tusimple} and outperforms the state-of-the-art techniques \cite{tabelini2021,zheng2021} on a new dataset, called SDLane, containing structurally more diverse lanes.

This work has the following major contributions:
\begin{itemize}
\itemsep0mm
\item We propose the notion of eigenlanes, which are data-driven lane descriptors, to represent structurally diverse lanes compactly in the eigenlane space.
\item We develop SIIC-Net to detect and regress road lanes in the eigenlane space effectively and efficiently. It yields outstanding performances on various datasets.
\item We construct the SDLane dataset to represent structurally diverse and complicated lanes in real driving environments more faithfully than the existing datasets do.\footnote{SDLane is available at \href{https://www.42dot.ai/akit/dataset}{https://www.42dot.ai/akit/dataset}.}
\end{itemize}

\section{Related Work}

In autonomous driving systems, it is required to detect boundaries of road lanes, sidewalks, or crosswalks precisely and reliably. Whereas early methods \cite{he2004,aly2008,hillel2014,zhou2010} adopted hand-crafted low-level features,
several CNN-based lane detectors have been developed recently to cope with complicated road scenes using deep features. Most of these techniques are based on the semantic segmentation framework \cite{pan2018,neven2018,ghafoorian2018gan,hou2019_road,hou2020_inter,zheng2021}, in which pixel-wise classification is performed to decide whether each pixel belongs to a lane or not.
In \cite{pan2018}, Pan \etal developed a convolutional network to propagate spatial information between pixels through message passing. Zheng \etal \cite{zheng2021} passed the information more efficiently using a recurrent feature aggregation module. In \cite{hou2019_road}, Hou \etal proposed a self-attention distillation mechanism to train the network more effectively. Also, Hou \etal \cite{hou2020_inter} employed teacher and student networks to transfer structural relationships between lanes by constructing an inter-region affinity graph. To maintain long-range consistency of segmented results, Ghafoorian \etal \cite{ghafoorian2018gan} used a discriminator to refine prediction results of a generator through adversarial training. In \cite{neven2018}, Neven \etal applied a perspective transformation to segmented pixels of each lane and used the transformed points for polynomial fitting.

Alternative approaches, different from the segmentation framework, also have been developed. In \cite{philion2019}, a network predicts the probability that vertically neighboring pixels belong to the same lane. Then, through greedy iterations, trajectories of pixels are concatenated to form a full lane. In \cite{wang2020poly}, a three-branched network regresses polynomial coefficients of each lane and estimates its starting and ending points. In \cite{qin2020}, for computational efficiency, a network selects the location of each lane on a predefined set of rows only. In \cite{li2020curvelane}, a unified network blends multi-scale features and combines prediction results at different levels. Qu \etal \cite{qu2021} estimated multiple keypoints and associated them to reconstruct actual lanes.
Meanwhile, an anchor-based detection framework was employed for lane detection \cite{li2019line,tabelini2021}.
These anchor-based techniques consider straight lines as lane candidates (or anchors) and generate a predefined set of candidates. Then, they classify and regress each candidate by estimating the lane probability and offset vectors. Despite providing promising results, they may fail to detect highly curved lanes. The proposed algorithm is also anchor-based but can deal with such complicated lanes successfully by employing eigenlane descriptors.

\begin{figure}[t]

  \centering
  \includegraphics[width=1\linewidth]{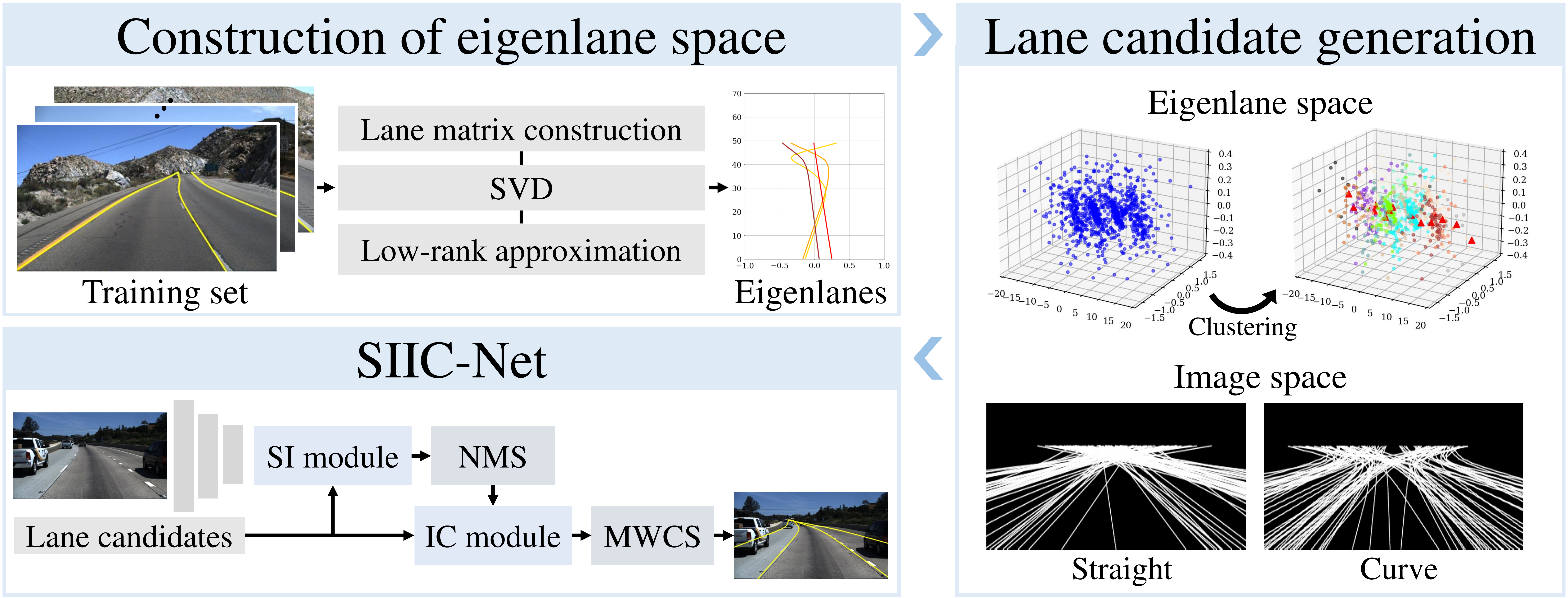}
  \caption{Overview of the proposed algorithm. It is recommended to watch the accompanying video of the proposed algorithm.}
  \label{fig:Overview_fig}
\end{figure}

\section{Proposed Algorithm}
We propose a novel algorithm to detect structurally diverse road lanes in the eigenlane space. Figure \ref{fig:Overview_fig} shows an overview of the proposed algorithm.
First, the eigenlane space is constructed by performing the low-rank approximation of lanes in a training set. Second, lane candidates are generated by clustering lanes in the eigenlane space. Third, given an image, an optimal set of lanes are determined from the lane candidates by SIIC-Net.


\subsection{Eigenlanes -- Formulation}
\label{ssec:s-net}

SVD and principal component analysis (PCA) are used in various fields to represent data compactly \cite{Linear2012,y2015_SVD_Hopcroft}. A well-known such application is face recognition using eigenfaces \cite{turk1991face}. Also, in this conference, eigencontours \cite{park2022} are proposed to describe object boundaries. In this paper, we use SVD to represent road lanes. Specifically, we adopt a data-driven approach and exploit the distribution of lanes in a training set, instead of employing parametric curves such as polynomials \cite{savitzky1964} or splines \cite{gordon1974,bartels1995}, to represent lanes.

\vspace*{0.1cm}
\noindent\textbf{Definition of eigenlane space:} A lane can be represented by 2D points sampled uniformly in the vertical direction. Specifically, let $\bfx=[x_1, x_2, \ldots, x_N]^\top$ be a lane, where $x_i$ is the $x$-coordinate of the $i$th sample and $N$ is the number of samples. We construct a lane matrix ${\mathbf A}=[{\mathbf x}_1, {\mathbf x}_2, \cdots, {\mathbf x}_L]$ from a training set containing $L$ lanes. Then, we apply SVD to the lane matrix $\mathbf A$ by
\begin{equation}\label{eq:svd}
    \textstyle
    \mathbf{A} = \mathbf{U} \mathbf{\Sigma} \mathbf{V}^\top
\end{equation}
where $\mathbf{U} = [\bfu_1, \cdots, \bfu_N]$ and $\bfV = [\bfv_1, \cdots, \bfv_L]$ are orthogonal matrices and $\mathbf{\Sigma}$ is a diagonal matrix, composed of singular values $\sigma_1 \geq \sigma_2 \geq \cdots \geq \sigma_r > 0$. Here, $r$ is the rank of $\bfA$. It is known that
\begin{equation}\label{eq:rank-m}
    \bfA_M  = [\tilde{\bfx}_1, \cdots, \tilde{\bfx}_L] = \sigma_{1}{\mathbf u}_{1}{\mathbf v}^\top_{1} + \cdots + \sigma_{M}{\mathbf u}_{M}{\mathbf v}^\top_{M}
\end{equation}
is the best rank-$M$ approximation of $\bfA$ in that the Frobenius norm $\|\bfA - \bfA_M\|_F$ is minimized \cite{y2015_SVD_Hopcroft}. Also, the sum of squared lane approximation errors is given by
\begin{equation} \label{eq:approx_error}
\textstyle
\|\bfA - \bfA_M\|_F^2 = \sum_{i=1}^L \|\bfx_i - \tilde{\bfx}_i\|^2 = \sum_{i=M+1}^r \sigma_i^2.
\end{equation}

\begin{figure}[t]
  \centering
  \includegraphics[width=1\linewidth]{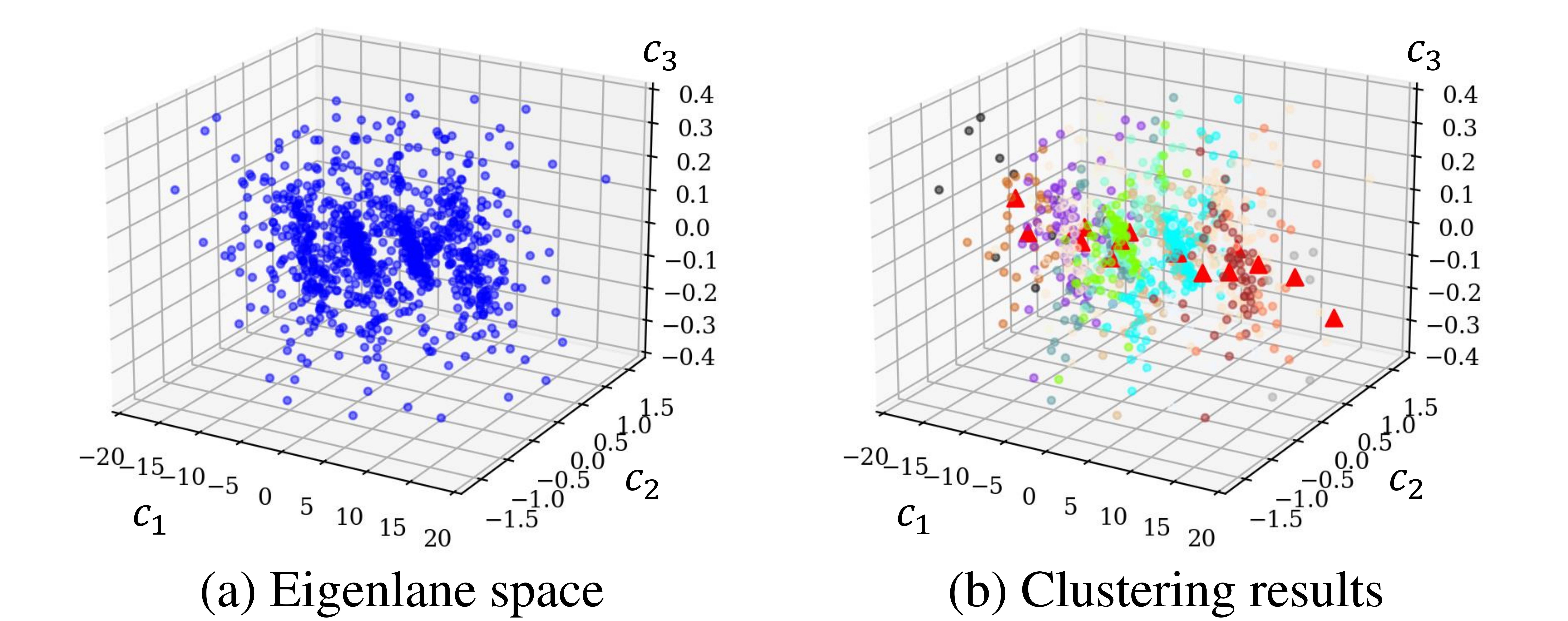}
  \caption{(a) 1,000 training lanes sampled from the TuSimple dataset \cite{tusimple} are visualized in the 3D eigenlane space. (b) These lanes are clustered using the $K$-means algorithm with $K=16$. Each cluster is in a different color, and the centroids are depicted by red triangles.}
  \label{fig:EigenCluster}
\end{figure}

In \eqref{eq:rank-m}, each approximate lane $\tilde{\bfx}_i$ is given by a linear combination of the first $M$ left singular vectors $\bfu_1, \cdots, \bfu_M$. In other words,
\begin{equation} \label{eq:x_approx}
\tilde{\bfx}_i = \bfU_M \bfc_i = [\bfu_1, \cdots, \bfu_M] \bfc_i.
\end{equation}
We refer to these $\bfu_1, \cdots, \bfu_M$ as \textit{eigenlanes}, because they are eigenvectors of $\bfA \bfA^\top$. They can be regarded as principal components in PCA. However, strictly speaking, this is not PCA, since the mean lane is not removed in constructing $\bfA$ \cite{Linear2012}. We do not remove the mean (or center the data) because we are interested in the best low-rank approximation, instead of finding the best fitting subspace \cite{y2015_SVD_Hopcroft}.

We call the space spanned by $\{\bfu_1, \cdots, \bfu_M\}$ as the \textit{eigenlane space}. Given a lane $\bfx$, we project it onto the eigenlane space to obtain the approximation
\begin{equation}\label{eq:backward}
    \tilde{\bfx} = \bfU_M \bfc
\end{equation}
where the coefficient vector $\bfc$ is given by
\begin{equation}\label{eq:forward}
    \bfc = \bfU_M^\top \bfx.
\end{equation}
Thus, in the eigenlane space, a lane $\bfx$ is approximately represented by the $M$-dimensional vector $\bfc$ in \eqref{eq:forward}. Also, the approximate $\bfx$ can be reconstructed from $\bfc$ via \eqref{eq:backward}.

\begin{algorithm}[t]
\caption{Lane candidate generation in eigenlane space}
    {\bf Input:} Set of training lanes $\{\bfx_1, \bfx_2, \cdots, \bfx_L\}$, $M=$ \# of eigenlanes, $K=$ \# of lane candidates
    \begin{algorithmic}[1]
        \State Construct the lane matrix ${\mathbf A}$ and perform SVD in \eqref{eq:svd}
        \State Transform each lane $\bfx_i$ to $\bfc_i$ via \eqref{eq:forward}
        \State Apply the $K$-means algorithm \cite{hartigan1979} to $\{\bfc_1, \bfc_2, \cdots, \bfc_L\}$ to obtain $K$ centroids $\bfc^{1}, \cdots, \bfc^{K}$.
        \State Generate the lane candidate $\bfl_k = \bfU_M \bfc^k$ by inversely transforming each centroid $\bfc^{k}$ via \eqref{eq:backward}
    \end{algorithmic}
    {\bf Output:} Set of lane candidates $\{ \bfl_{1}, \cdots, \bfl_{K} \}$
    \label{alg:DRC_algorithm}
\end{algorithm}

\vspace*{0.1cm}
\noindent\textbf{Detection and regression in eigenlane space:} Let $\{\tilde{\bfx}_1, \ldots, \tilde{\bfx}_L\}$ be the set of training lanes, which are already approximated via \eqref{eq:x_approx}. By clustering these lanes, we obtain a finite number of lane candidates (or anchors) for lane detection. However, instead of the original lane space of dimension $N$, we perform the clustering in the eigenlane space of dimension $M$, as illustrated in Figure~\ref{fig:EigenCluster}. This is possible because the transform $\bfU_M$ is length-preserving;
\begin{equation}\label{eq:dist}
\| \tilde{\bfx}_i - \tilde{\bfx}_j\| =  \|\bfc_i - \bfc_j\|.
\end{equation}
Also, $M < N$. The clustering in the lower-dimensional space is more effective and more efficient. {\bf Algorithm 1} summarizes the process of lane candidate generation.

Suppose that a lane candidate $\bfl = \bfU_M \bfc$ is detected. Then, we refine it to
\begin{equation}\label{eq:reg}
     \bfl + \Delta \bfl = \bfU_M (\bfc + \Delta \bfc),
\end{equation}
by finding an offset vector $\Delta \bfc$ using a regressor. This is also done in the eigenlane space, since $\|\Delta \bfl\| = \|\Delta \bfc\|$.

\subsection{Eigenlanes -- Image examples}
\label{ssec:eigenlanes_examples}

\begin{figure}[t]
  \centering
  \includegraphics[width=1\linewidth]{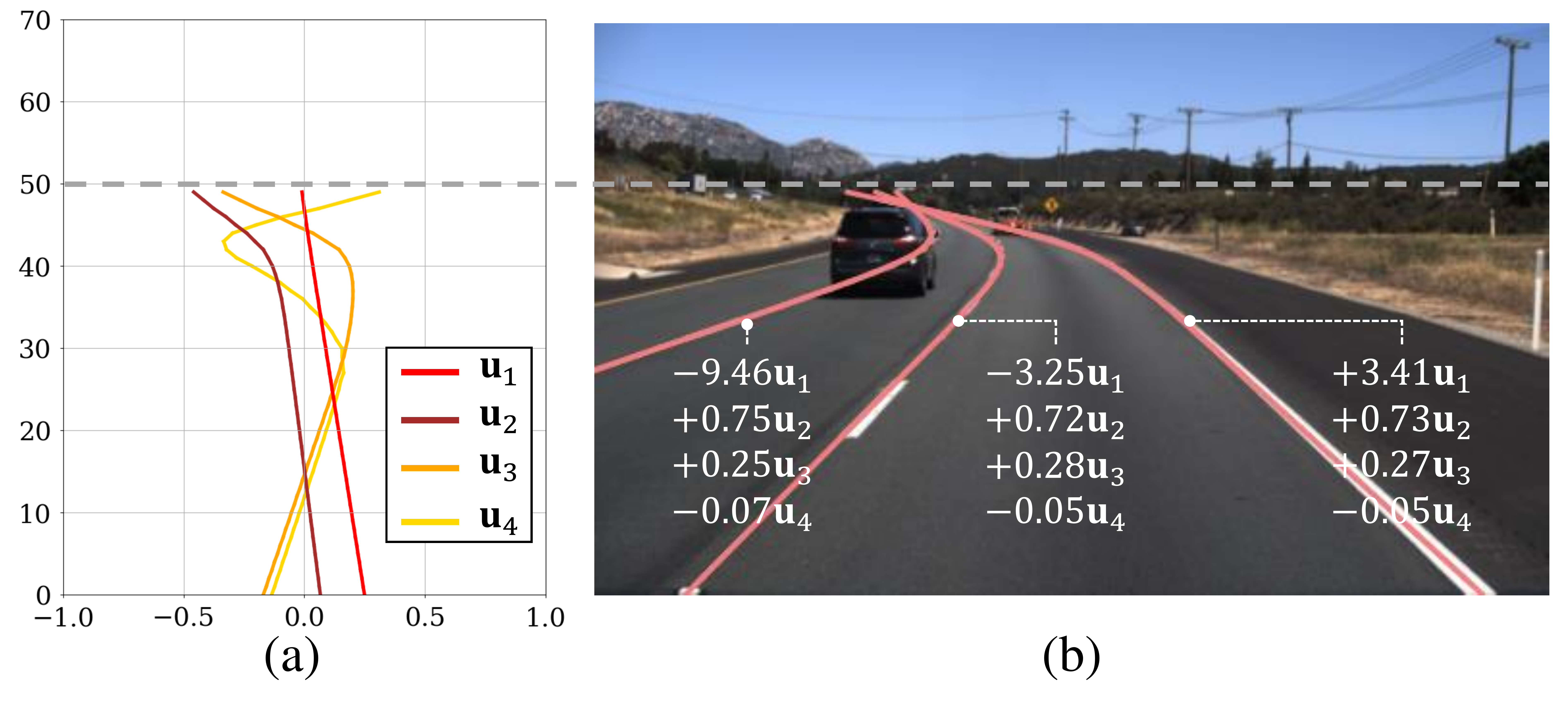}
  \vspace*{-0.8cm}
  \caption{(a) The first four eigenlanes $\bfu_1, \bfu_2, \bfu_3, \bfu_4$ for the TuSimple dataset. (b) Three example lanes are approximated by linear combinations of the four eigenlanes. }
  \label{fig:Eigenspace_fig}
\end{figure}

\noindent\textbf{Eigenlanes in image space:} In this example, we use the TuSimple dataset \cite{tusimple} to determine eigenlanes. Here, each lane is represented by a 50D vector, \ie $N=50$.  Figure~\ref{fig:Eigenspace_fig}(a) shows the four eigenlanes $\bfu_1, \bfu_2, \bfu_3, \bfu_4$, which are sufficient to represent all TuSimple lanes faithfully. The first eigenlane $\bfu_1$ is a slanted line, instead of a vertical line. Most lanes are slanted from the viewpoint of a driving car. Because $\bfu_1$ achieves the best rank-1 approximation of these lanes, it is also slanted. By weighting $\bfu_1$, we can represent straight road lanes in an image. Next, $\bfu_2$ is slightly curved at the top side (far from the cameras), and $\bfu_3$ is curvier. These eigenlanes are required to represent simply curved lanes. Finally, $\bfu_4$ has an inflection point and is used to describe highly complicated lanes.

In Figure~\ref{fig:Eigenspace_fig}(b), the straight line parts of the left and middle lanes are slanted to the right, whereas $\bfu_1$ is slanted to the left. Thus, the coefficients for $\bfu_1$ for these two lanes are negative. For all lanes, the coefficients for $\bfu_2$ and $\bfu_3$ are not negligible because the lanes are curved. They are, however, not complicated, so their 4th coefficients are insignificant.

\begin{figure}[t]
\vspace{-0.3cm}
    \subfloat[Original lanes] {\includegraphics[width=2.05cm,height=1.4cm]{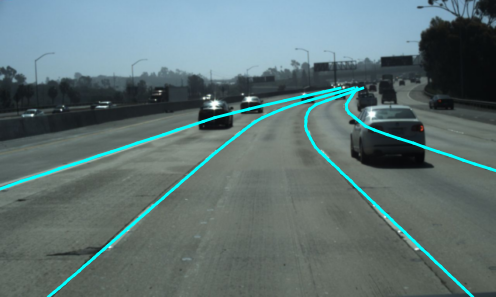}}\,\!\!
    \subfloat[$M=1$] {\includegraphics[width=2.05cm,height=1.4cm]{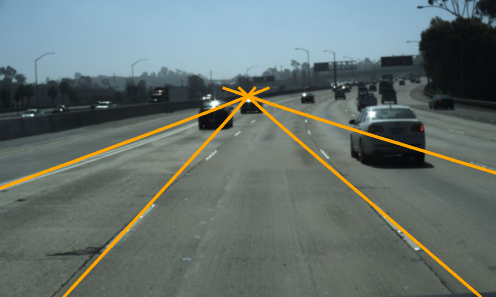}}\,\!\!
    \subfloat[$M=2$] {\includegraphics[width=2.05cm,height=1.4cm]{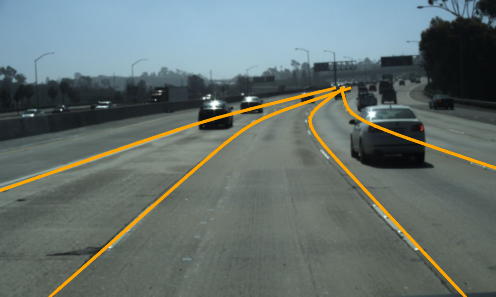}}\,\!\!
    \subfloat[$M=3$] {\includegraphics[width=2.05cm,height=1.4cm]{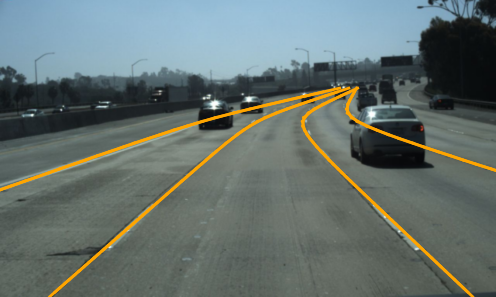}}\\
    \vspace*{-0.2cm}
    \caption
    {
        Rank-$M$ approximation: original lanes in (a) are reconstructed by the first (b) one, (c) two, and (d) three eigenlanes, respectively.
    }
    \label{fig:Approx_fig}
\end{figure}

\vspace*{0.1cm}
\noindent\textbf{Rank-$M$ approximation:}
Figure~\ref{fig:Approx_fig} shows examples of original lanes and their rank-$M$ approximations. In (b), using only one eigenlane $\bfu_1$, the rank-1 approximation yields line parts of the lanes. In (c), the rank-2 approximation reconstructs curved parts additionally. To represent the curved parts more faithfully, the rank-3 approximation in (d) is required, which matches well the ground-truth in (a).

\begin{figure}[t]
\vspace{-0.3cm}
    \subfloat {\raisebox{1.5em}{\rotatebox[origin=t]{90}{\scriptsize TuSimple}}}\hspace{-0.001cm}\,\!\!
    \subfloat {\includegraphics[width=2.00cm,height=1.4cm]{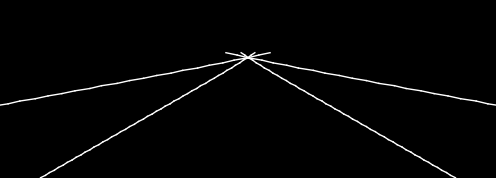}}\,\!\!
    \subfloat {\includegraphics[width=2.00cm,height=1.4cm]{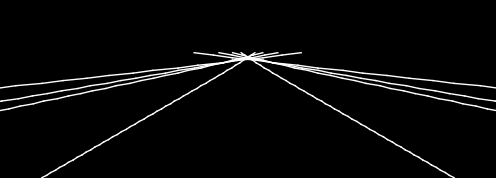}}\,\!\!
    \subfloat {\includegraphics[width=2.00cm,height=1.4cm]{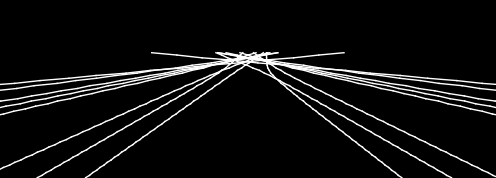}}\,\!\!
    \subfloat {\includegraphics[width=2.00cm,height=1.4cm]{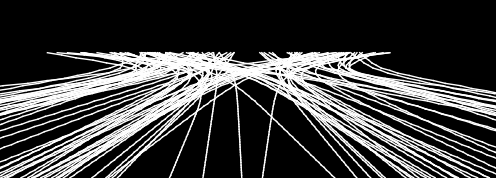}}\\[-4.7ex]

    \subfloat {\raisebox{1.5em}{\rotatebox[origin=t]{90}{\scriptsize CULane}}}\hspace{-0.01cm}\,\!
    \subfloat {\includegraphics[width=2.00cm,height=1.4cm]{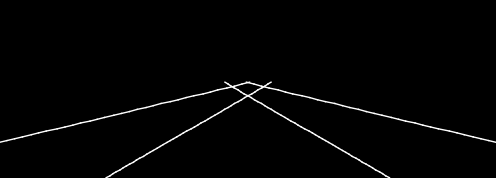}}\,\!\!
    \subfloat {\includegraphics[width=2.00cm,height=1.4cm]{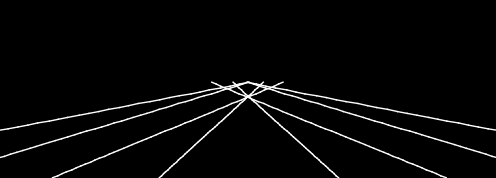}}\,\!\!
    \subfloat {\includegraphics[width=2.00cm,height=1.4cm]{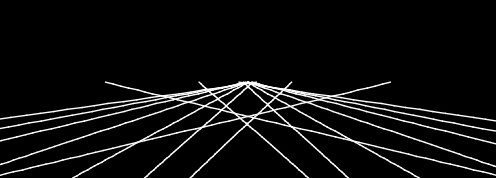}}\,\!\!
    \subfloat {\includegraphics[width=2.00cm,height=1.4cm]{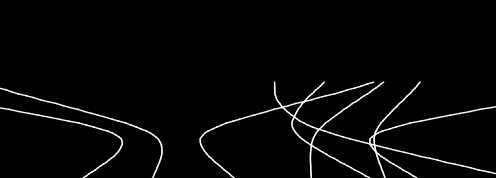}}\\[-4.7ex]

    \setcounter{subfigure}{-1}

    \subfloat {\raisebox{1.5em}{\rotatebox[origin=t]{90}{\scriptsize SDLane}}}\hspace{-0.01cm}\,\!
    \subfloat[$K$=4] {\includegraphics[width=2.00cm,height=1.4cm]{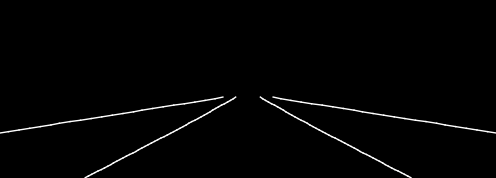}}\,\!\!
    \subfloat[$K$=8] {\includegraphics[width=2.00cm,height=1.4cm]{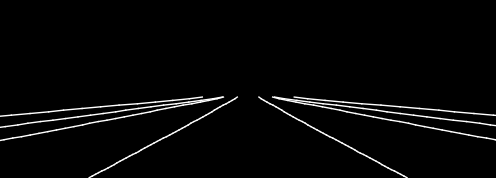}}\,\!\!
    \subfloat[$K$=16] {\includegraphics[width=2.00cm,height=1.4cm]{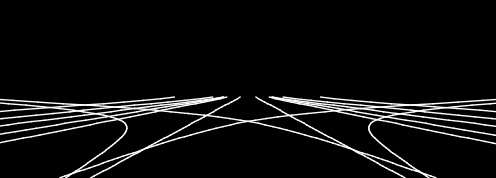}}\,\!\!
    \subfloat[$K$=500] {\includegraphics[width=2.00cm,height=1.4cm]{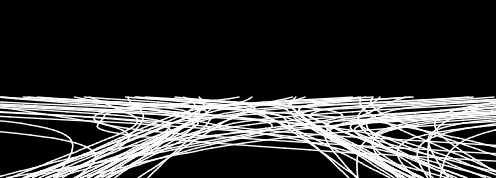}}\\[-1.7ex]

    \caption
    {
        Lane candidates, generated by the $K$-means clustering, for the TuSimple \cite{tusimple}, CULane \cite{pan2018}, and SDLane datasets. In (d), only curved lanes are shown among 500 candidates. In CULane, there are only 8 curved lanes among those 500 candidates.
    }
    \label{fig:Candidates_fig}
\end{figure}

\vspace*{0.1cm}
\noindent\textbf{Lane candidates:} As mentioned previously, we generate lane candidates for detection, by grouping training lanes in the eigenlane space using the $K$-means algorithm. Figure~\ref{fig:Candidates_fig} shows such generated candidates according to $K$. Being the centroids, they are representative of all training lanes. Notice that the proposed SDLane dataset contains many curved lanes with high curvatures, whereas the existing CULane dataset \cite{pan2018} consists of mainly straight lanes. TuSimple contains curved lanes, which, however, lack diversity.

\subsection{SIIC-Net}
\begin{figure*}[t]
\vspace{-0.2cm}
  \centering
  \includegraphics[width=1\linewidth]{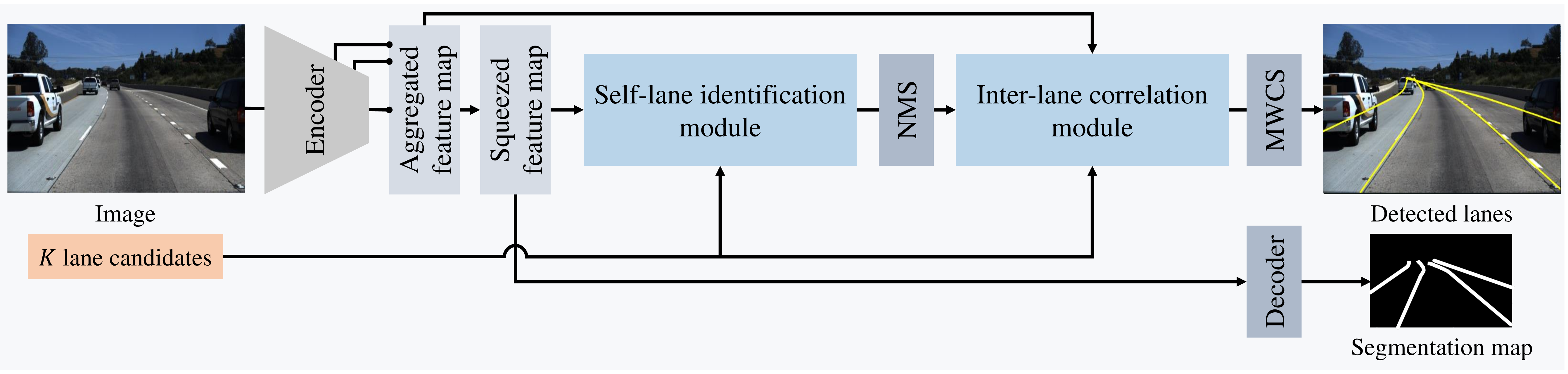}
  \caption{The architecture of the proposed SIIC-Net: Given an image, the encoder extracts two types of feature maps and the decoder yields a segmentation map. Then, the self-lane identification module (SI) and the inter-lane correlation (IC) module process the feature maps. After the SI module, NMS removes redundant lane candidates. After the IC module, MWCS determines an optimal set of lanes.}
  \vspace{-0.2cm}
  \label{fig:Network_fig}
\end{figure*}

Using the $K$ lane candidates $\{ \bfl_1, \cdots, \bfl_K \}$, the proposed SIIC-Net detects road lanes. In Figure~\ref{fig:Network_fig}, SIIC-Net consists of the encoder-decoder part, SI module, and IC module. After the SI module, the non-maximum suppression (NMS) is performed to filter out redundant candidates. Also, after the IC module, the maximum weight clique selection (MWCS) is done to determine an optimal set of lanes.

\vspace*{0.1cm}
\noindent\textbf{Encoder-decoder part:}
We adopt ResNet50 \cite{he2016deep} as the encoder to extract features and employ the auxiliary branch \cite{qin2020} as the decoder to yield a binary segmentation map of lanes. From an image, we extract multi-scale feature maps and aggregate the three lowest-level maps. To this end, we match the resolutions of the two smaller maps to the finest one via bilinear interpolation. Let $X_{\rm a}=[X_{\rm a}^1, X_{\rm a}^2, \ldots, X_{\rm a}^{C_1}]\in \mathbb{R}^{H\times W\times C_1}$ be the aggregated feature map, where $H$, $W$, and $C_1$ are the feature height, the feature width, and the number of channels. Then, we squeeze $X_{\rm a}$ using convolutional layers to yield $X_{\rm s}\in \mathbb{R}^{H\times W\times C_2}$. The decoder processes $X_{\rm s}$ to produce the segmentation map. We use the decoder part in the training phase only, as in \cite{qin2020}.

\begin{figure}[t]
  \centering
  \includegraphics[width=1\linewidth]{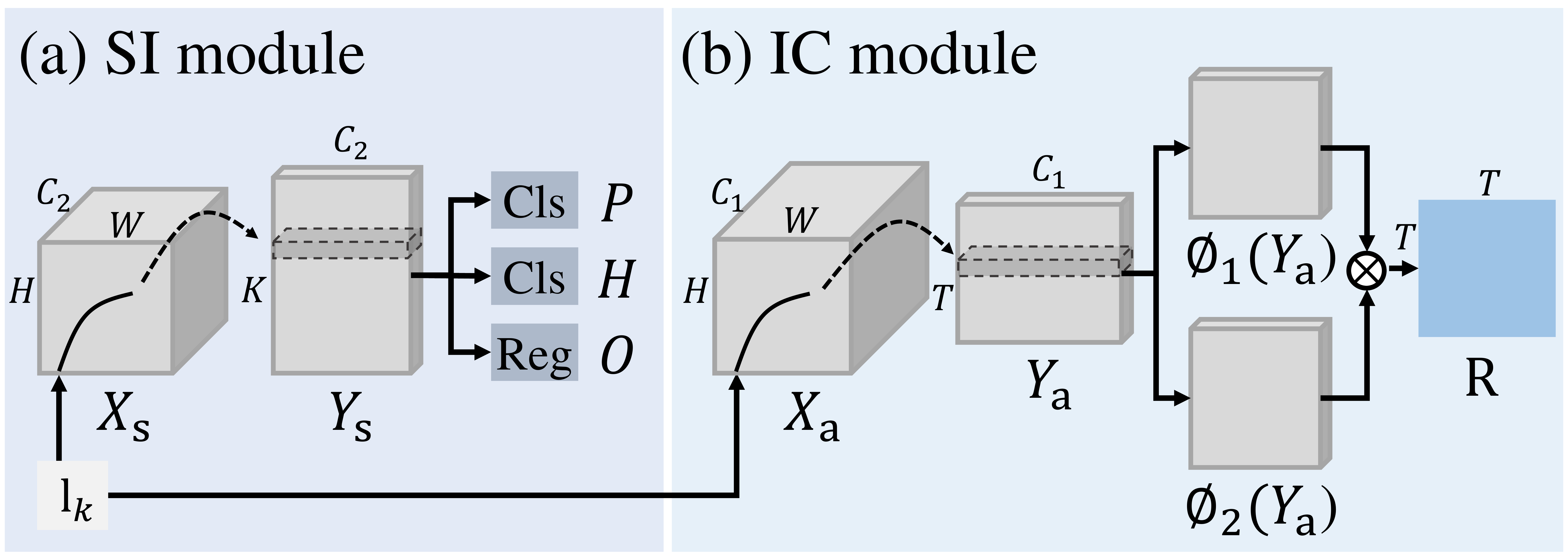}
  \caption{Block diagrams of the SI and IC modules.}
  \vspace{-0.3cm}
  \label{fig:Module_fig}
\end{figure}

\vspace*{0.1cm}
\noindent\textbf{Self-lane identification (SI) module:}
For each lane candidate $\bfl_k$, we estimate the lane probability, the positional offset, and the height of the topmost point using the SI module, the structure of which is in  Figure~\ref{fig:Module_fig}(a). The SI module employs a line pooling layer \cite{lee2017, han2020}. From the squeezed feature map $X_{\rm s}$, it obtains the lane feature map $Y_{\rm s}=[Y_{\rm s}^1, Y_{\rm s}^2, \ldots, Y_{\rm s}^{C_2}]\in \mathbb{R}^{K \times C_2}$ by averaging the features of pixels along $\mathbf{l}_k$;
\begin{equation}\label{eq:line_pooling}
    \textstyle
    {Y_{\rm s}^c({k})} = \frac{1}{|\mathbf{l}_k|} \sum_{\mathbf{p} \in {\mathbf{l}_k}}{X_{\rm s}^c}(\mathbf{p})
\end{equation}
for $1\leq k \leq K$ and $1 \leq c \leq C_2$, where $|\mathbf{l}_k|$ denotes the number of pixels in $\mathbf{l}_k$. Then, two probability vectors and a lane offset matrix are obtained by
\begin{equation}
    \textstyle
    P = \sigma(f_1(Y)), \;\;\; H = \sigma(f_2(Y)), \;\;\;    O = f_3(Y)
\end{equation}
where $f_1$ and $f_2$ are fully-connected layers of sizes $C_2\times 2$ and $C_2\times R$ for classification, $f_3$ is a fully-connected layer of size $C_2\times M$ for regression, and $\sigma(\cdot)$ is the softmax function. For lane candidate $\bfl_{k}$, $P_{k}$ informs the probabilities that it is a lane or not, $H_{k}$ represents the probabilities that its ending point is located at one of $R$ pre-defined heights, and $O_{k}= \Delta \mathbf{c}_k$ is an offset vector in \eqref{eq:reg} for lane refinement.

\begin{figure}[t]
\vspace{-0.3cm}
\setlength{\belowcaptionskip}{-0.5cm}
    \begin{flushright}
    \setcounter{subfigure}{0}
    \subfloat[(a)] {\includegraphics[width=2.05cm,height=1.4cm]{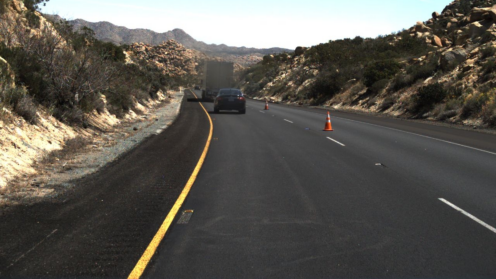}}\,\!\!
    \subfloat[(b)] {\includegraphics[width=2.05cm,height=1.4cm]{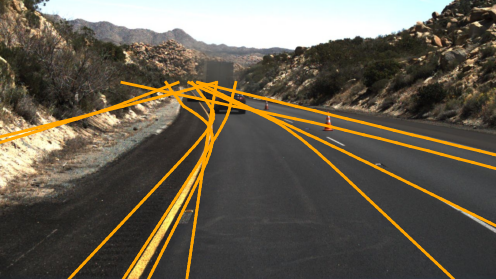}}\,\!\!
    \subfloat[(c)] {\includegraphics[width=2.05cm,height=1.4cm]{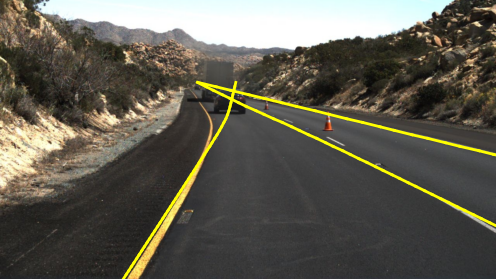}}\,\!\!
    \subfloat[(d)] {\includegraphics[width=2.05cm,height=1.4cm]{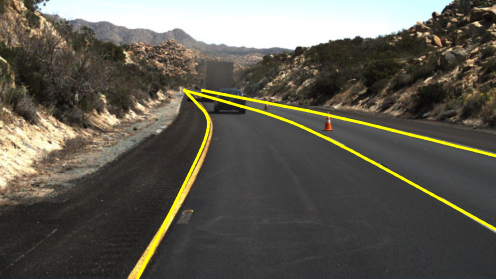}}\\
    \vspace{-0.2cm}
    \caption
    {
        An example of the lane detection by SIIC-Net: (a) input image, (b) 10 selected lanes after NMS, (c) optimal lanes determined by MWCS, and (d) refined lanes using the regression offsets from the SI module.
    }
    \label{fig:SIIC_result}
    \end{flushright}
\end{figure}

\vspace*{0.1cm}
\noindent\textbf{NMS:}
Many redundant lanes tend to be detected around an actual one. We filter out those overlapping ones through an NMS process after the SI module. We select the most reliable lane $\mathbf{l}_{i^\star}$ by
\begin{equation}
    \textstyle
    i^\star =  \arg \max_{i} P_i.
\end{equation}
Then, we remove overlapping lanes, whose the intersection over union (IoU) ratios with the selected lane are higher than a threshold. We perform this process $T$ times to select the $T$ reliable lanes. The default $T$ is 10. Note that we focus on reducing false negatives, rather than false positives. Figure~\ref{fig:SIIC_result}(b) shows 10 selected lanes after the NMS process.

\vspace*{0.1cm}
\noindent\textbf{Inter-lane correlation (IC) module:} In general, adjacent lanes are equally distanced in road environments. Also, under the perspective projection, lanes intersect at a vanishing point in a 2D image. Because of these structural constraints, lanes are highly correlated with one another. To exploit this correlation, we design the IC module, which estimates the relation score between every pair of selected lanes. Figure~\ref{fig:Module_fig}(b) shows the structure of the IC module.

Given the aggregated feature map $X_{\rm a}$, IC yields a lane feature map $Y_{\rm a}$ similarly to \eqref{eq:line_pooling}. $Y_{\rm a}$ is a $T\times C_1$ matrix, in which each row contains the $C_1$-dimensional feature vector for a selected lane. Then, it obtains the relation matrix
\begin{equation}
    \textstyle
    {\mathbf R}=\phi_1(Y_{\rm a})\times \phi_2(Y_{\rm a})^\top
\end{equation}
of size $T\times T$. Here, $\phi_1$ and $\phi_2$ are feature transforms, implemented by convolution layers and the $l_2$-normalization. Thus, each element of $\mathbf R$ is a score in $[-1, 1]$, representing how compatible the corresponding pair of lanes are.

\captionsetup[subfigure]{labelformat=empty}
\captionsetup[subfigure]{labelformat=parens}

\noindent\textbf{MWCS:}
We determine an optimal set of lanes by employing MWCS \cite{jin2021}, which is a graph optimization technique. We first construct a complete graph $G=({\cal V}, {\cal E})$. The node set ${\cal V}=\{v_1, v_2, \ldots, v_{T}\}$ represents the  $T$ selected lanes from NMS. Every pair of lanes are connected by an edge in the edge set ${\cal E}=\{(v_i, v_j): i \neq j\}$. Each edge is assigned weight $w(v_i, v_j)=\frac{{\mathbf R}(i, j) + {\mathbf R}(j, i)}{2}$.

Let $\theta$ denote a clique, represented by the index set of member nodes. We define the compatibility $E_{\rm compatible}(\theta)$ of clique $\theta$ as
\begin{equation}
    \textstyle
    E_{\rm compatible}(\theta) = \sum_{i \in \theta} \sum_{j \in \theta, j>i} w(v_i, v_j).
\end{equation}
We then select the maximal weight clique $\theta^{\star}$ by
\begin{equation}\label{eq:maximal_clique}
    \textstyle
    \theta^{\star} = \arg \max_{\theta} E_{\rm compatible}(\theta)
\end{equation}
subject to a constraint $w(v_i, v_j)>\kappa$ for all edges in the clique, where $\kappa$ is a threshold. If there is no clique satisfying the constraint, we select the maximal single-node clique $\theta^{\star} = \{ i^{\star} \}$, where $i^{\star} =  \arg \max_{i} P_{v_i}$.

Next, we refine each lane in the optimal clique $\theta^{\star}$ by $\mathbf{U}(\mathbf{c}_{v_i} + \Delta\mathbf{c}_{v_i})$, where $\Delta\mathbf{c}_{v_i}$ is the offset vector in the eigenlane space, predicted by the SI module. Moreover, we refine the vertical height of each lane, by removing sampled points whose $y$-coordinates are bigger than $H_{v_i}$. Figure~\ref{fig:SIIC_result}(c) and (d) show the MWCS results and their refined ones.

The supplemental document (Section A) describes the training process and the architecture of SIIC-Net in detail.

\section{Experimental Results}

\subsection{Datasets}
\label{ssec:dataset}

\noindent\textbf{TuSimple \cite{tusimple}:}
It consists of 6,408 images only, which are split into 3,268 training, 358 validation, and 2,782 test images. For each image, lanes are annotated by the 2D coordinates of sampling points with a uniform height interval of 10 pixels. It contains both straight and curved lanes, whose shapes are, however, simple and similar to one another.

\vspace*{0.1cm}
\noindent\textbf{CULane \cite{pan2018}:} It is a rich dataset with about 130K images. Its 34,680 test images are classified into 9 categories. In some categories, lanes are highly implied or even invisible. For each image, pixel-wise masks for up to 4 road lanes are provided. Most lanes are straight lines.

\begin{figure}[t]
\vspace{-0.3cm}
    \subfloat {\includegraphics[width=2.05cm,height=1.2cm]{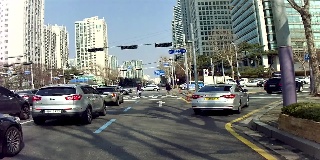}}\,\!\!
    \subfloat {\includegraphics[width=2.05cm,height=1.2cm]{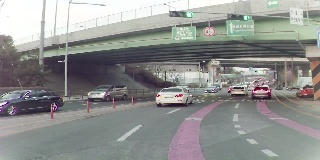}}\,\!\!
    \subfloat {\includegraphics[width=2.05cm,height=1.2cm]{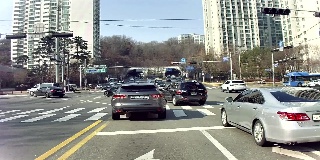}}\,\!\!
    \subfloat {\includegraphics[width=2.05cm,height=1.2cm]{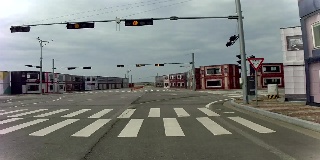}}\\[-4.7ex]

    \subfloat {\includegraphics[width=2.05cm,height=1.2cm]{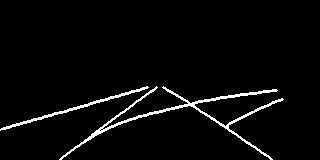}}\,\!\!
    \subfloat {\includegraphics[width=2.05cm,height=1.2cm]{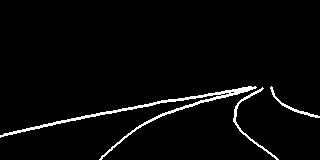}}\,\!\!
    \subfloat {\includegraphics[width=2.05cm,height=1.2cm]{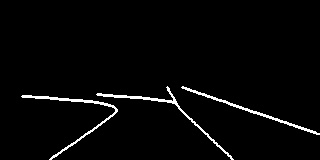}}\,\!\!
    \subfloat {\includegraphics[width=2.05cm,height=1.2cm]{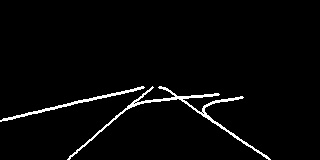}}\\[-2.7ex]

    \caption
    {
        Example images and ground-truth lanes in the SDLane dataset. Since crossroad images are included in SDLane, some lanes for left or right turns are highly curved and implicit.
    }
    \label{fig:SDLane}
\end{figure}

\vspace*{0.1cm}
\noindent\textbf{SDLane:} We construct a structurally diverse lane dataset SDLane. It contains highly curved and complicated lanes, as illustrated in Figure~\ref{fig:SDLane}. We collect 43K images, which are split into about 39K training and 4K testing images, and annotate the actual lanes manually. As mentioned in Section~\ref{ssec:eigenlanes_examples} and Figure~\ref{fig:Candidates_fig}, SDLane contains more curved lanes than CULane and more diverse lanes than TuSimple.
The structural diversity of SDLane is discussed in detail in the supplemental document (Section B).

\subsection{Evaluation metrics}

In TuSimple, a lane point is regarded as correctly located if its distance to the ground-truth point is shorter than a threshold \cite{tusimple}. Then, the accuracy is defined as $\frac{N_{\rm c}}{N}$, where $N$ is the number of ground-truth lane points, and $N_{\rm c}$ is the number of correctly predicted lane points. Also, the false positive rate (FPR) and the false negative rate (FNR) are
\begin{equation}\label{eq:ts_err}
    \textstyle
    {\rm FPR} = \frac{F_{\rm pred}}{N_{\rm pred}}, \;\;\; {\rm FNR} = \frac{M_{\rm pred}}{N_{\rm gt}}
\end{equation}
where $F_{\rm pred}$ is the number of incorrectly predicted lanes, $N_{\rm pred}$ is that of predicted lanes, $M_{\rm pred}$ is that of missed lanes, and $N_{\rm gt}$ is that of ground-truth lanes.

In CULane and SDLane, each lane is regarded as a thin stripe with 30 pixel width \cite{pan2018}. A predicted lane is declared to be correct if its IoU ratio with the ground-truth is greater than 0.5. The precision and the recall are computed by
\begin{equation}\label{eq:pre_rec}
    \textstyle
    {\rm Precision} = \frac{\rm TP}{\rm TP + FP}, \;\;\; {\rm Recall} = \frac{\rm TP}{\rm TP + FN}
\end{equation}
where $\rm TP$ is the number of correctly detected lanes, $\rm FP$ is that of false positives, and $\rm FN$ is that of false negatives. Then, the F-measure is computed by
\begin{equation}\label{eq:fscore}
    \textstyle
    {\rm F \text{-}\rm measure} =  \frac{2 \times \rm Precision \times \rm Recall}{\rm Precision + \rm Recall}.
\end{equation}

\begin{table}[t]\centering
    \renewcommand{\arraystretch}{0.8}
    \caption
    {
        Comparison on TuSimple. Only the algorithms with publicly available source codes are compared.
    }
    \vspace*{-0.15cm}
    \resizebox{0.85\linewidth}{!}{
    \begin{tabular}[t]{+L{2.6cm}^C{1.5cm}^C{1.5cm}^C{1.5cm}}
    \toprule
    & Accuracy & FPR & FNR\\
    \midrule
         LaneNet \cite{neven2018}        & 96.38 & 0.0780 & 0.0244\\
         SCNN \cite{pan2018}             & 96.53 & 0.0617 & \bf{0.0180}\\
         SAD \cite{hou2019_road}         & \underline{96.64} & 0.0602 & \underline{0.0205}\\
         UFast \cite{qin2020}            & 95.82 & 0.1905 & 0.0392 \\
         RESA \cite{zheng2021}           & \bf{96.82} & 0.0363 & 0.0248 \\
         LaneATT \cite{tabelini2021}     & 95.63 & \underline{0.0353} & 0.0292 \\
    \midrule
         Proposed                   & 95.62 & \bf{0.0320} & 0.0399\\
    \bottomrule
    \end{tabular}}
    \label{table:tusimple}
\end{table}

\captionsetup[subfigure]{labelformat=empty}
\begin{figure}[t]
\vspace{-0.2cm}
\setlength{\belowcaptionskip}{-0.3cm}
    \begin{flushright}

    \subfloat {\includegraphics[width=2.05cm,height=1.2cm]{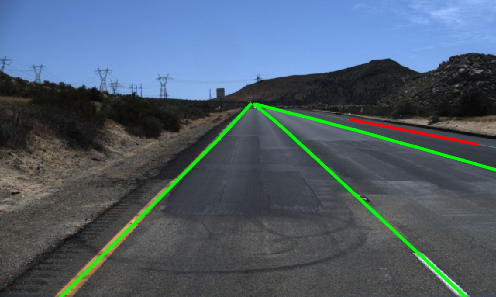}}\,\!\!
    \subfloat {\includegraphics[width=2.05cm,height=1.2cm]{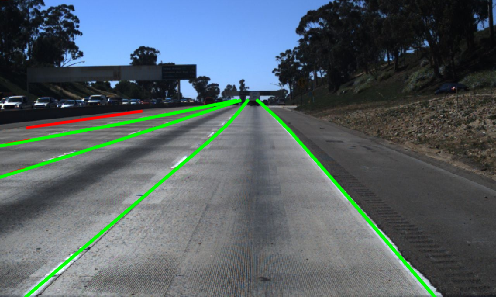}}\,\!\!
    \subfloat {\includegraphics[width=2.05cm,height=1.2cm]{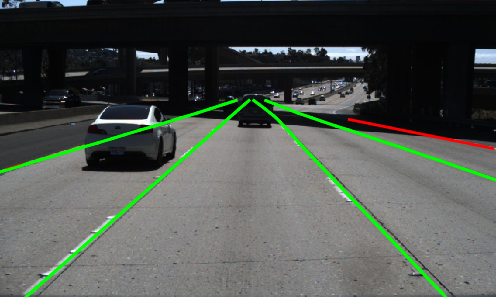}}\,\!\!
    \subfloat {\includegraphics[width=2.05cm,height=1.2cm]{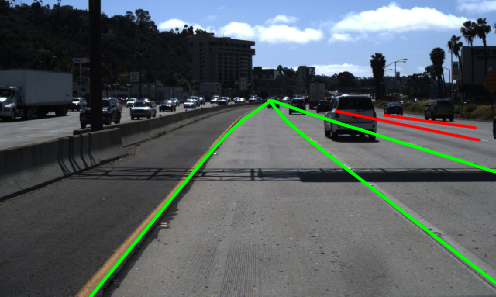}}
    \caption
    {
        Detection results of the proposed algorithm on the TuSimple dataset. Detected lanes are depicted in green, while false negatives are in red.
    }
    \label{fig:tusimple_result}
    \end{flushright}
    \vspace*{-0.1cm}
\end{figure}
\captionsetup[subfigure]{labelformat=parens}

\begin{table*}[t]\centering
\vspace*{-0.1cm}
    \renewcommand{\arraystretch}{0.8}

    \caption
    {
        Comparison of the F\text{-}{\rm measure} performances (\%) on the CULane dataset, whose lanes are classified into 9 categories. For the `Crossroad' category, only $\rm FP$ is reported. * means that the encoder backbone is ResNet18.
    }
    \vspace*{-0.15cm}
    \resizebox{0.95\linewidth}{!}{
    \begin{tabular}[t]{+L{2.5cm}^C{1.2cm}^C{1.2cm}^C{1.2cm}^C{1.2cm}^C{1.2cm}^C{1.2cm}^C{1.2cm}^C{1.2cm}^C{1.2cm}^C{1.2cm}}
    \toprule
    Category & Normal & Crowded & Night & No line & Shadow & Arrow & Dazzle & Curve & Crossroad & Total\\
    \midrule
         SCNN \cite{pan2018}              & 90.6 & 69.7 & 66.1 & 43.4 & 66.9 & 84.1 & 58.5 & 65.7 & 1990 & 71.6\\
         SAD \cite{hou2019_road}          & 90.7 & 70.0 & 66.3 & 43.5 & 67.0 & 84.4 & 59.9 & 65.7 & 2052 & 71.8\\
         UFast \cite{qin2020}               & 90.7 & 70.2 & 66.7 & 44.4 & 69.3 & 85.7 & 59.5 & \underline{69.7} & 2037 & 72.3\\
         Curve-Nas \cite{li2020curvelane}& 90.7 & 72.3 & 68.9 & 49.4 & 70.1 & 85.8 & 67.7 & 68.4 & 1746 & 74.8\\
         RESA \cite{zheng2021}          & \bf{92.1} & 73.1 & 69.9 & 47.7 & 72.8 & \bf{88.3} & 69.2 & \bf{70.3} & 1503 & 75.3\\
         LaneATT* \cite{tabelini2021}    & 91.1 & 73.0 & 69.0 & 48.4 & 70.9 & 85.5 & 65.7 & 63.4 & \bf{1170} & 75.1\\
         LaneATT \cite{tabelini2021}     & \underline{91.7} & \bf{76.2} & 70.8 & 50.5 & \bf{76.3} & 86.3 & 69.5 & 64.1 & \underline{1264} & \underline{77.0}\\
    \midrule
         Proposed*                        & 91.5 & 74.8 & \underline{71.4} & \underline{51.1} & 72.3 & \underline{87.7} & \underline{69.7} & 62.0 & 1507 & 76.5\\
         Proposed                         & \underline{91.7} & \underline{76.0} & \bf{71.8} & \bf{52.2} & \underline{74.1} & \underline{87.7} & \bf{69.8} & 62.9 & 1509 & \bf{77.2}\\
    \bottomrule
    \end{tabular}}
    \label{table:culane}
    \vspace*{-0.1cm}
\end{table*}

\subsection{Comparative assessment}

\vspace*{0.1cm}
\noindent\textbf{Comparison on TuSimple:}
Table~\ref{table:tusimple} compares the proposed algorithm with the conventional road lane detectors \cite{neven2018,pan2018,hou2019_road,qin2020,li2019line} on TuSimple. The proposed algorithm yields a high FNR, resulting in a relatively low accuracy, but provides the best FPR performance. Figure~\ref{fig:tusimple_result} shows some detection results. Most errors are caused by the lanes, which are far from the camera and thus short. Except for them, the proposed algorithm detects most lanes precisely, especially ego and alternative lanes, which are more important for driving.

\begin{figure}[t]
\vspace{-0.2cm}
\setlength{\belowcaptionskip}{-0.4cm}
    \begin{flushright}

    \subfloat {\raisebox{1.15em}{\rotatebox[origin=t]{90}{\scriptsize Crowded}}}\hspace{-0.01cm}\,\!
    \subfloat {\includegraphics[width=2.00cm,height=1.1cm]{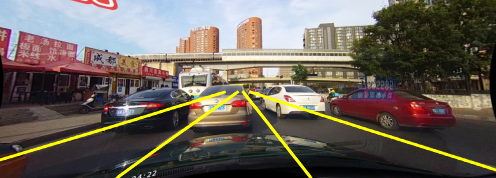}}\,\!\!
    \subfloat {\includegraphics[width=2.00cm,height=1.1cm]{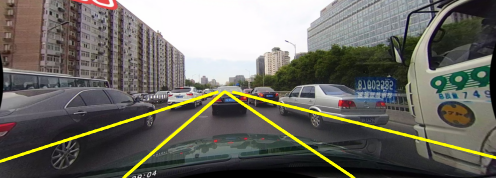}}\,\!\!
    \subfloat {\includegraphics[width=2.00cm,height=1.1cm]{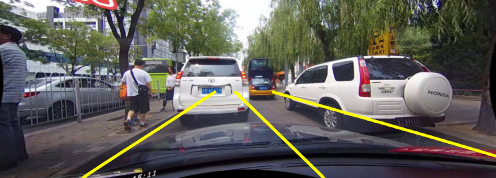}}\,\!\!
    \subfloat {\includegraphics[width=2.00cm,height=1.1cm]{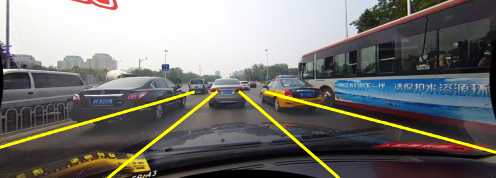}}\\[-2.2ex]

    \subfloat {\raisebox{1.15em}{\rotatebox[origin=t]{90}{\scriptsize Night}}}\hspace{-0.01cm}\,\!\!
    \subfloat {\includegraphics[width=2.00cm,height=1.1cm]{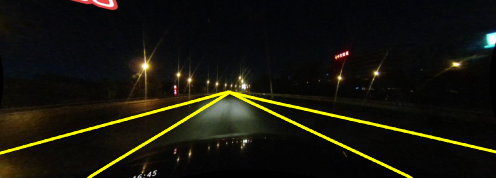}}\,\!\!
    \subfloat {\includegraphics[width=2.00cm,height=1.1cm]{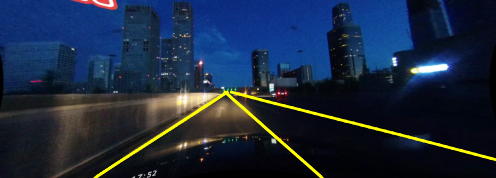}}\,\!\!
    \subfloat {\includegraphics[width=2.00cm,height=1.1cm]{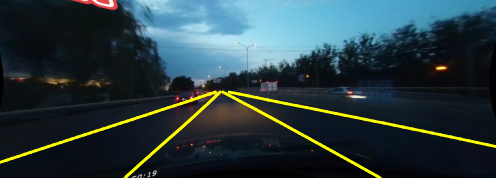}}\,\!\!
    \subfloat {\includegraphics[width=2.00cm,height=1.1cm]{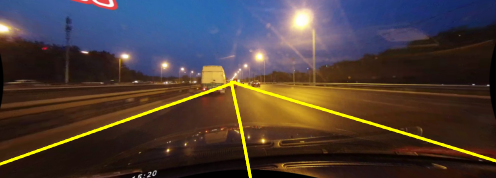}}\\[-2.2ex]

    \subfloat {\raisebox{1.15em}{\rotatebox[origin=t]{90}{\scriptsize No line}}}\hspace{-0.01cm}\,\!
    \subfloat {\includegraphics[width=2.00cm,height=1.1cm]{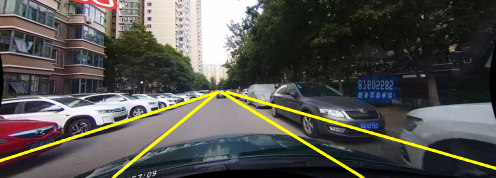}}\,\!\!
    \subfloat {\includegraphics[width=2.00cm,height=1.1cm]{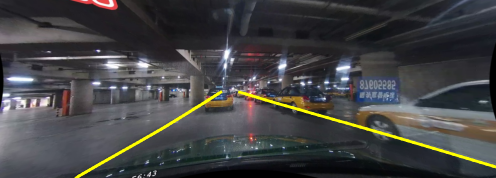}}\,\!\!
    \subfloat {\includegraphics[width=2.00cm,height=1.1cm]{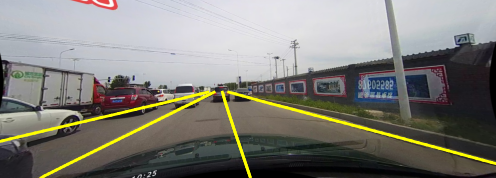}}\,\!\!
    \subfloat {\includegraphics[width=2.00cm,height=1.1cm]{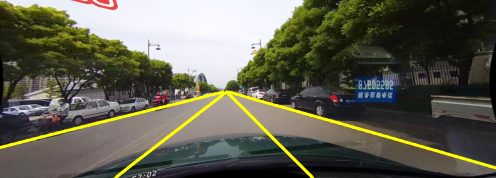}}\\[-2.2ex]

    \subfloat {\raisebox{1.15em}{\rotatebox[origin=t]{90}{\scriptsize Shadow}}}\hspace{-0.01cm}\,\!
    \subfloat {\includegraphics[width=2.00cm,height=1.1cm]{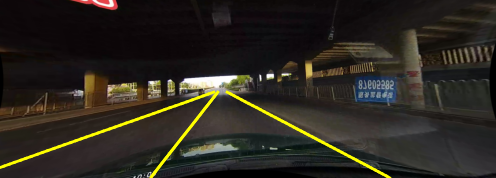}}\,\!\!
    \subfloat {\includegraphics[width=2.00cm,height=1.1cm]{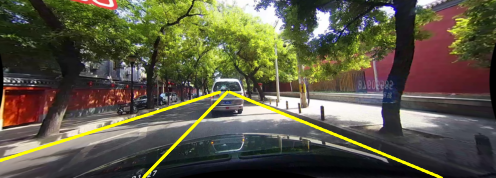}}\,\!\!
    \subfloat {\includegraphics[width=2.00cm,height=1.1cm]{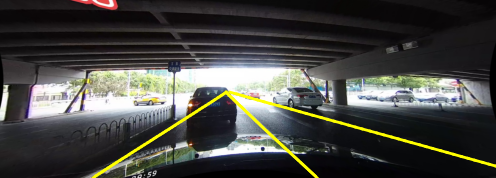}}\,\!\!
    \subfloat {\includegraphics[width=2.00cm,height=1.1cm]{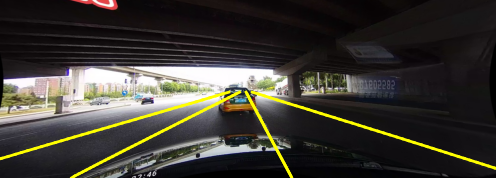}}\\[-2.2ex]

    \subfloat {\raisebox{1.15em}{\rotatebox[origin=t]{90}{\scriptsize Dazzle}}}\hspace{-0.01cm}\,\!
    \subfloat {\includegraphics[width=2.00cm,height=1.1cm]{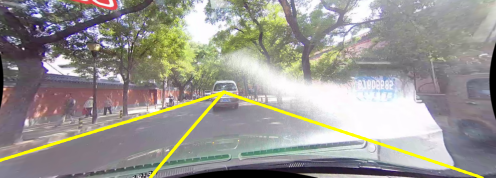}}\,\!\!
    \subfloat {\includegraphics[width=2.00cm,height=1.1cm]{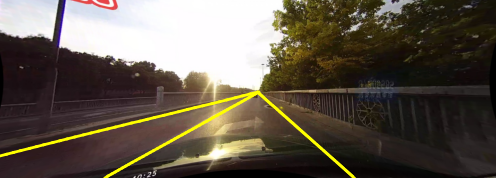}}\,\!\!
    \subfloat {\includegraphics[width=2.00cm,height=1.1cm]{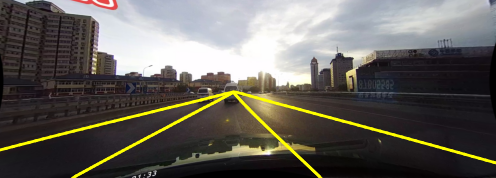}}\,\!\!
    \subfloat {\includegraphics[width=2.00cm,height=1.1cm]{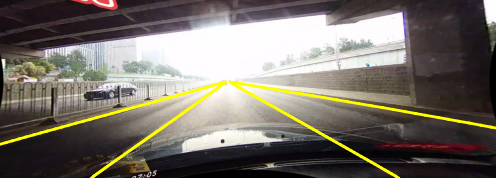}}\\[-0.5ex]

    \caption
    {
        Detection results of the proposed algorithm on five challenging categories in the CULane dataset.
    }
    \label{fig:culane_result}
    \end{flushright}
    \vspace*{-0.1cm}
\end{figure}

\vspace*{0.1cm}
\noindent\textbf{Comparison on CULane:}
Table~\ref{table:culane} compares the F-measure performances on CULane, whose lanes are divided into 9 categories. The proposed algorithm outperforms all conventional algorithms. Especially, the proposed algorithm yields excellent results on the challenging categories of `Night,' `No line,' and `Dazzle' in which lanes are highly implicit or even invisible. This indicates that the proposed algorithm can detect challenging lanes by considering the correlation or compatibility among detected lanes. The performance gap against LaneATT \cite{tabelini2021} is marginal. But, when the same backbone of ResNet18 is used for both algorithms, the gap increases further. On the `Curve' category, the proposed algorithm is inferior to the other methods. However, it does not mean that the proposed algorithm is not capable of detecting curved lanes. Since the proportion of curved lanes is only 1.2\%, the CULane training data are not enough to generate curved lane candidates in the eigenlane space. Figure~\ref{fig:culane_result} shows some detection results in challenging scenarios. Although the lanes are extremely ambiguous, the proposed algorithm detects them reliably by exploiting the structural properties between the adjacent ones.

\captionsetup[subfigure]{labelformat=empty}
\captionsetup[subfigure]{labelformat=parens}

\vspace*{0.1cm}
\noindent\textbf{Comparison on SDLane:}
It is challenging to detect highly curved lanes in the anchor-based detection framework, but the proposed algorithm provides excellent results on such curved lanes. To demonstrate this, on SDLane, we compare the proposed algorithm with the state-of-the-art techniques \cite{tabelini2021,zheng2021,liu2021}. LaneATT \cite{tabelini2021} is an anchor-based method considering straight lines as anchors, while RESA \cite{zheng2021} is based on the semantic segmentation framework. Recently, CondLaneNet\cite{liu2021} was proposed, which yields an F-measure of 79.48\% on CULane. We train these methods on SDLane using the publicly available source codes.

\begin{figure*}[t]
\vspace{-0.3cm}
\setlength{\belowcaptionskip}{-0.4cm}
    \begin{flushright}

    \subfloat {\raisebox{1.4em}{\rotatebox[origin=t]{90}{\scriptsize Image}}}\,\!\hspace{-0.03cm}
    \subfloat {\includegraphics[width=2.42cm,height=1.3cm]{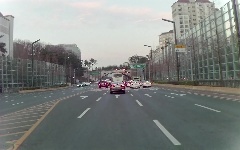}}\,\!\!
    \subfloat {\includegraphics[width=2.42cm,height=1.3cm]{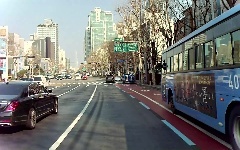}}\,\!\!
    \subfloat {\includegraphics[width=2.42cm,height=1.3cm]{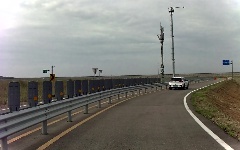}}\,\!\!
    \subfloat {\includegraphics[width=2.42cm,height=1.3cm]{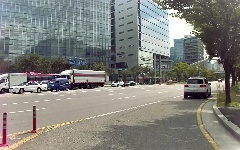}}\,\!\!
    \subfloat {\includegraphics[width=2.42cm,height=1.3cm]{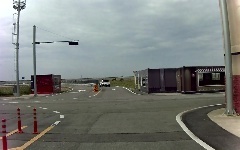}}\,\!\!
    \subfloat {\includegraphics[width=2.42cm,height=1.3cm]{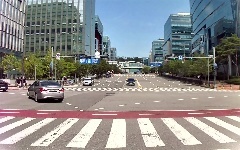}}\,\!\!
    \subfloat {\includegraphics[width=2.42cm,height=1.3cm]{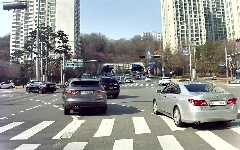}}\\[-2.2ex]

    \subfloat {\raisebox{1.4em}{\rotatebox[origin=t]{90}{\scriptsize LaneATT \cite{tabelini2021}}}}\,\!
    \subfloat {\includegraphics[width=2.42cm,height=1.3cm]{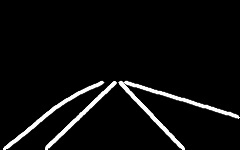}}\,\!\!
    \subfloat {\includegraphics[width=2.42cm,height=1.3cm]{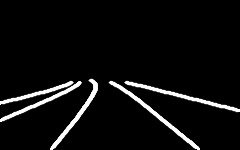}}\,\!\!
    \subfloat {\includegraphics[width=2.42cm,height=1.3cm]{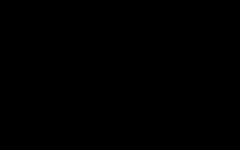}}\,\!\!
    \subfloat {\includegraphics[width=2.42cm,height=1.3cm]{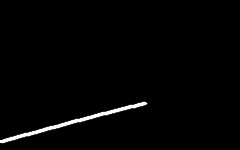}}\,\!\!
    \subfloat {\includegraphics[width=2.42cm,height=1.3cm]{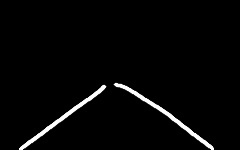}}\,\!\!
    \subfloat {\includegraphics[width=2.42cm,height=1.3cm]{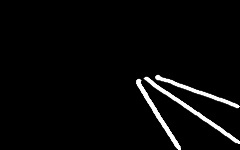}}\,\!\!
    \subfloat {\includegraphics[width=2.42cm,height=1.3cm]{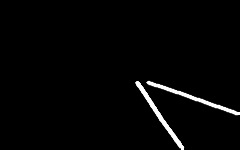}}\\[-2.45ex]

    \subfloat {\raisebox{1.4em}{\rotatebox[origin=t]{90}{\scriptsize RESA \cite{zheng2021}}}}\,\!
    \subfloat {\includegraphics[width=2.42cm,height=1.3cm]{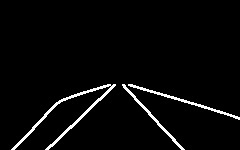}}\,\!\!
    \subfloat {\includegraphics[width=2.42cm,height=1.3cm]{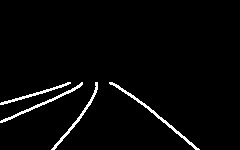}}\,\!\!
    \subfloat {\includegraphics[width=2.42cm,height=1.3cm]{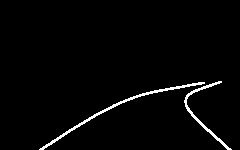}}\,\!\!
    \subfloat {\includegraphics[width=2.42cm,height=1.3cm]{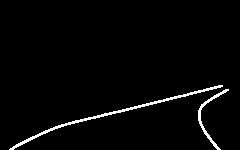}}\,\!\!
    \subfloat {\includegraphics[width=2.42cm,height=1.3cm]{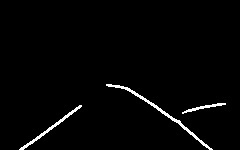}}\,\!\!
    \subfloat {\includegraphics[width=2.42cm,height=1.3cm]{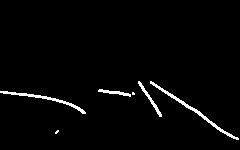}}\,\!\!
    \subfloat {\includegraphics[width=2.42cm,height=1.3cm]{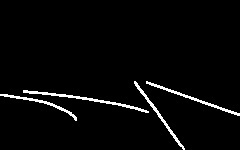}}\\[-2.2ex]

    \setcounter{subfigure}{-1}

    \subfloat {\raisebox{1.4em}{\rotatebox[origin=t]{90}{\scriptsize Proposed}}}\,\!
    \subfloat {\includegraphics[width=2.42cm,height=1.3cm]{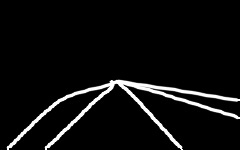}}\,\!\!
    \subfloat {\includegraphics[width=2.42cm,height=1.3cm]{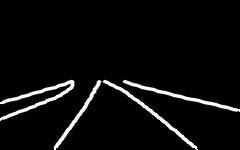}}\,\!\!
    \subfloat {\includegraphics[width=2.42cm,height=1.3cm]{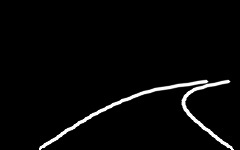}}\,\!\!
    \subfloat {\includegraphics[width=2.42cm,height=1.3cm]{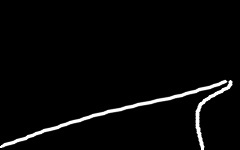}}\,\!\!
    \subfloat {\includegraphics[width=2.42cm,height=1.3cm]{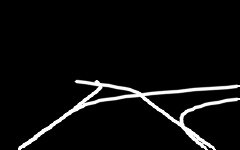}}\,\!\!
    \subfloat {\includegraphics[width=2.42cm,height=1.3cm]{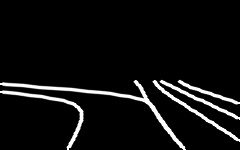}}\,\!\!
    \subfloat {\includegraphics[width=2.42cm,height=1.3cm]{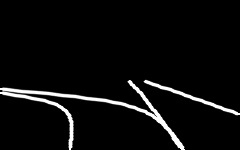}}\\[-2.2ex]

    \setcounter{subfigure}{-1}

    \subfloat {\raisebox{1.4em}{\rotatebox[origin=t]{90}{\scriptsize Ground-truth}}}\,
    \subfloat {\includegraphics[width=2.42cm,height=1.3cm]{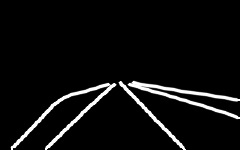}}\,\!\!
    \subfloat {\includegraphics[width=2.42cm,height=1.3cm]{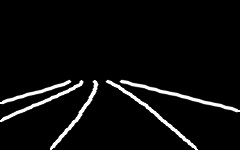}}\,\!\!
    \subfloat {\includegraphics[width=2.42cm,height=1.3cm]{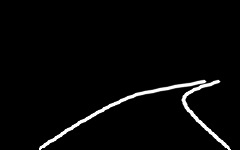}}\,\!\!
    \subfloat {\includegraphics[width=2.42cm,height=1.3cm]{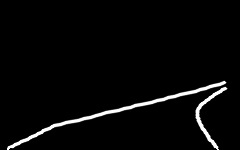}}\,\!\!
    \subfloat {\includegraphics[width=2.42cm,height=1.3cm]{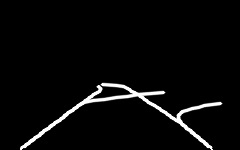}}\,\!\!
    \subfloat {\includegraphics[width=2.42cm,height=1.3cm]{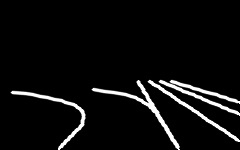}}\,\!\!
    \subfloat {\includegraphics[width=2.42cm,height=1.3cm]{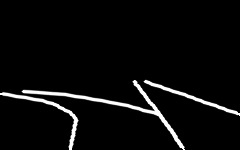}}\\[-0.5ex]

    \caption
    {
        Comparison of lane detection results on the SDLane dataset.
    }
    \label{fig:sdlane_result}
    \end{flushright}
    \vspace*{-0.1cm}
\end{figure*}

In Table~\ref{table:sdlane}, we see that the proposed algorithm is superior to the existing methods. LaneATT poorly recalls highly curved lanes, because straight anchors deviate too much from such lanes. Although RESA yields a higher recall rate, it does not detect invisible lanes reliably. CondLaneNet achieves the highest precision score, but its recall rate is still low.
Figure~\ref{fig:sdlane_result} compares some detection results. LaneATT detects straight or mildly curved lanes precisely, but it fails to detect more complicated lanes, even though those lanes are visible in the images. RESA detects such complicated lanes better than LaneATT does. However, for invisible or unobvious lanes, it does not preserve the continuous lane structure in detection results. In contrast, the proposed algorithm is capable of detecting both straight and curved lanes precisely, as well as processing implicit lanes reliably. This is because the proposed algorithm generates diverse lane candidates and then localizes lanes effectively in the eigenlane space.

%
%

The proposed algorithm yields an F-measure of 80.47\% on SDLane, which is significantly higher than that on the `Curve' category in CULane in Table~\ref{table:culane}. This means that, with sufficiently big data, the proposed algorithm can deal with curved lanes effectively. More results are presented in the supplemental document (Section C) and video.

\begin{table}[t]\centering
    \renewcommand{\arraystretch}{0.8}
    \caption
    {
        Comparison on SDLane.
    }
    \vspace*{-0.15cm}
    \resizebox{0.85\linewidth}{!}{
    \begin{tabular}[t]{+L{2.9cm}^C{1.4cm}^C{1.4cm}^C{1.5cm}}
    \toprule
    & Precision & Recall & F-measure\\
    \midrule
         LaneATT \cite{tabelini2021}       & 85.78 & 64.28 & 73.49\\
         RESA \cite{zheng2021}       & 82.35 & \underline{72.46} & \underline{77.09}\\
         CondLaneNet \cite{liu2021}       & \bf{87.59} & 67.08 & 75.97\\
         Proposed                   & \underline{86.04} & \bf{75.58} & \bf{80.47}\\
    \bottomrule
    \end{tabular}}
    \label{table:sdlane}
    \vspace*{-0.1cm}
\end{table}

\captionsetup[subfigure]{labelformat=empty}
\captionsetup[subfigure]{labelformat=parens}

\subsection{Ablation studies}
We conduct ablation studies to analyze the efficacy of eigenlane projection and SIIC-Net components on the SDLane dataset. Also, we analyze the runtime for each component of SIIC-Net.

\vspace*{0.1cm}
\noindent\textbf{Efficacy of eigenlane projection:}
In the anchor-based detection framework, it is important to generate a set of candidates (anchors) reliably.
Table~\ref{table:ablation_anchor} compares alternative methods for lane candidate generation. Method \RomNum{1} (proposed) clusters 1,000 lane candidates in the eigenlane space of dimension $M=6$, while \RomNum{2} does 1,000 candidates in the original lane space of $N=50$. \RomNum{3} and \RomNum{4}, respectively, obtain 1,000 and 10,000 straight lines as done in \cite{tabelini2021}. The mean intersection over union (mIoU) scores are reported. For each lane in test data, the closest candidate in terms of IoU is found. Then, the average of these matching IoU's is computed. Between \RomNum{1} and \RomNum{2}, the scores are similar. However, method \RomNum{1} can refine lane candidates more efficiently using compact offset vectors ($M<N$). \RomNum{3} yields poor results. By requiring 10 times more candidates, \RomNum{4} performs better than \RomNum{3} does, but the gap with \RomNum{1} is still high. This is because \RomNum{3} and \RomNum{4} do not consider curved lanes. In contrast, the proposed notion of eigenlanes enables the systematic generation of curved lane candidates.

\begin{table}[t]\centering
    \renewcommand{\arraystretch}{0.88}
    \caption
    {
        Ablation studies for the proposed eigenlane projection on the SDLane dataset. mIoU scores are reported.
    }
    \vspace*{-0.15cm}
    \resizebox{0.88\linewidth}{!}{
    \begin{tabular}[t]{+L{2.2cm}^C{1.4cm}^C{1.4cm}^C{1.4cm}^C{1.4cm}}
    \toprule
    & \RomNum{1} \scriptsize{(proposed)} & \RomNum{2} & \RomNum{3} & \RomNum{4}\\
    \midrule
         mIoU         & 0.814 & 0.815  & 0.691 & 0.738\\

    \bottomrule
    \end{tabular}}
    \label{table:ablation_anchor}
\end{table}

\begin{table}[t]\centering
    \renewcommand{\arraystretch}{0.95}
    \caption
    {
        Ablation studies for the components of SIIC-Net on the SDLane dataset.
    }
    \vspace*{-0.15cm}
    \resizebox{0.95\linewidth}{!}{
    \begin{tabular}[t]{+C{0.2cm}^L{3.5cm}^C{1.5cm}^C{1.5cm}^C{1.5cm}}
    \toprule
    & & Precision & Recall & F-measure\\
    \midrule
         \RomNum{1}. & w/o offsets         & 43.86 & 38.53  & 41.02\\
         \RomNum{2}. & w/o heights         & 80.27 & 70.52  & 75.08\\
         \RomNum{3}. & w/o IC+MWCS         & 82.98 & 74.82  & 78.69\\
         \RomNum{4}. & SIIC-Net            & 86.04 & 75.58 & 80.47\\
    \bottomrule
    \end{tabular}}
    \vspace*{-0.1cm}
    \label{table:ablation}
\end{table}

\vspace*{0.1cm}
\noindent\textbf{Efficacy of components in SIIC-Net:}
Table~\ref{table:ablation} compares several ablated methods. Method \RomNum{1} does not use regression offsets. In Method \RomNum{2}, height classification results are not used. Method \RomNum{3} uses the SI module and NMS only to detect road lanes. In Method \RomNum{3}, we modify NMS as follows. First, we stop the iteration if the probability is lower than 0.5. Second, we optimize the threshold for removing redundant lanes empirically.

Compared with the full SIIC-Net in \RomNum{4}, Method \RomNum{1} degrades performances severely, indicating that SIIC-Net estimates regression offsets accurately to refine detected lines. Also, from \RomNum{2} and \RomNum{4}, note that the height classification improves the performance by adjusting the ending points of lanes. Last, compared with \RomNum{4}, \RomNum{3} still yields lower performances in terms of all metrics even with those modifications of NMS. This means that the IC module with MWCS is required to detect lanes more precisely and more reliably.

\begin{table}[t]\centering
    \renewcommand{\arraystretch}{0.95}
    \caption
    {
        Analysis of running times of the proposed SIIC-Net. The processing times in seconds per frame are reported.
    }
    \vspace*{-0.15cm}
    \resizebox{0.95\linewidth}{!}{
    \begin{tabular}[t]{C{2.3cm}^C{2.3cm}^C{2.3cm}^C{2.3cm}}
    \toprule
    Encoding & SI+NMS & IC+MWCS & Total\\
    \midrule
     0.0036s & 0.0042s  & 0.0021s & 0.0099s\\
    \bottomrule
    \end{tabular}}
    \vspace*{-0.3cm}
    \label{table:runtime}
\end{table}

\vspace*{0.1cm}
\noindent\textbf{Runtime:} Table~\ref{table:runtime} shows the runtime for each stage of SIIC-Net. SI+NMS takes the longest time among the three stages, for it should perform feature pooling for all lane candidates. After the NMS process, IC+MWCS considers significantly fewer lanes, so it demands the lowest computational cost. Overall, the processing speed is about 101 frames per second, which is sufficiently fast for practical applications.

\section{Conclusions}
We proposed a novel algorithm to detect road lanes in the eigenlane space. First, we introduced the notion of eigenlanes, which are data-driven lane descriptors. Second, we generated a set of lane candidates by clustering training lanes in the eigenlane space. Third, we detected road lanes, by developing an anchor-based detection network SIIC-Net, from the lane candidates. Furthermore, we developed the structurally diverse dataset, containing highly curved and complicated lanes in real driving environments. Experimental results showed that the proposed algorithm provides excellent performances, especially on curved lanes.

\section*{Acknowledgements}

This work was supported by the National Research Foundation of Korea (NRF) grants funded by the Korea government (MSIT) (No.~NRF-2021R1A4A1031864 and No.~NRF-2022R1A2B5B03002310).

\clearpage

{\small
\bibliographystyle{ieee_fullname}
\bibliography{2022_CVPR_DKJIN}

\begin{thebibliography}{10}\itemsep=-1pt

\bibitem{tusimple}
{TuSimple} benchmark.
\newblock \url{https://github.com/TuSimple/tusimple-benchmark}.

\bibitem{aly2008}
Mohamed Aly.
\newblock Real time detection of lane markers in urban streets.
\newblock In {\em Intelligent Vehicles Symposium}, 2008.

\bibitem{bartels1995}
Richard~H. Bartels, John~C. Beatty, and Brian~A. Barsky.
\newblock {\em An Introduction to Splines for Use in Computer Graphics and
  Geometric Modeling}.
\newblock Morgan Kaufmann, 1995.

\bibitem{y2015_SVD_Hopcroft}
Avrim Blum, John Hopcroft, and Ravindran Kannan.
\newblock {Foundations of Data Science}.
\newblock 2015.

\bibitem{ghafoorian2018gan}
Mohsen Ghafoorian, Cedric Nugteren, N{\'o}ra Baka, Olaf Booij, and Michael
  Hofmann.
\newblock E{L}-{GAN}: {E}mbedding loss driven generative adversarial networks
  for lane detection.
\newblock In {\em ECCV Workshops}, 2018.

\bibitem{gordon1974}
William~J. Gordon and Richard~F. Riesenfeld.
\newblock B-spline curves and surfaces.
\newblock {\em Computer Aided Geometric Design}, pages 95--126, 1974.

\bibitem{han2020}
Qi Han, Kai Zhao, Jun Xu, and Ming-Ming Cheng.
\newblock Deep {H}ough transform for semantic line detection.
\newblock In {\em Proc. ECCV}, 2020.

\bibitem{hartigan1979}
J.~A. Hartigan and M.~A. Wong.
\newblock A {$K$}-means clustering algorithm.
\newblock {\em Journal of the Royal Statistical Society, Series C (Applied
  Statistics)}, 28(1):100--108, 1979.

\bibitem{he2016deep}
Kaiming He, Xiangyu Zhang, Shaoqing Ren, and Jian Sun.
\newblock Deep residual learning for image recognition.
\newblock In {\em Proc. IEEE CVPR}, 2016.

\bibitem{he2004}
Yinghua He, Hong Wang, and Bo Zhang.
\newblock Color-based road detection in urban traffic scenes.
\newblock {\em IEEE Trans. Intelligent Transportation Systems}, 5(4):309--318,
  2004.

\bibitem{hillel2014}
Aharon~Bar Hillel, Ronen Lerner, Dan Levi, and Guy Raz.
\newblock Recent progresss in road and lane detection{:} {A} survey.
\newblock {\em Mach. Vis. Appl.}, 25(3):727--745, 2014.

\bibitem{hou2020_inter}
Yuenan Hou, Zheng Ma, Chunxiao Liu, Tak-Wai Hui, and Chen~Change Loy.
\newblock Inter-region affinity distillation for road marking segmentation.
\newblock In {\em Proc. IEEE CVPR}, 2020.

\bibitem{hou2019_road}
Yuenan Hou, Zheng Ma, Chunxiao Liu, and Chen~Change Loy.
\newblock Learning lightweight lane detection {CNN}s by self attention
  distillation.
\newblock In {\em Proc. IEEE ICCV}, 2019.

\bibitem{jin2021}
Dongkwon Jin, Wonhui Park, Seong-Gyun Jeong, and Chang-Su Kim.
\newblock Harmonious semantic line detection via maximal weight clique
  selection.
\newblock In {\em Proc. IEEE CVPR}, 2021.

\bibitem{law2018cornernet}
Hei Law and Jia Deng.
\newblock Corner{N}et: {D}etecting objects as paired keypoints.
\newblock In {\em Proc. ECCV}, 2018.

\bibitem{Linear2012}
David~C. Lay.
\newblock {\em Linear Algebra and Its Applications}.
\newblock Pearson, 2007.

\bibitem{lee2017}
Jun-Tae Lee, Han-Ul Kim, Chul Lee, and Chang-Su Kim.
\newblock Semantic line detection and its applications.
\newblock In {\em Proc. IEEE ICCV}, 2017.

\bibitem{li2019line}
Xiang Li, Jun Li, Xiaolin Hu, and Jian Yang.
\newblock Line-{CNN}: {E}nd-to-end traffic line detection with line proposal
  unit.
\newblock {\em IEEE Trans. Intelligent Transportation Systems}, 21(1):248--258,
  2019.

\bibitem{liu2021}
Lizhe Liu, Xiaohao Chen, Siyu Zhu, and Ping Tan.
\newblock Cond{L}ane{N}et: {A} top-to-down lane detection framework based on
  conditional convolution.
\newblock In {\em Proc. IEEE ICCV}, 2021.

\bibitem{liu2016ssd}
Wei Liu, Dragomir Anguelov, Dumitru Erhan, Christian Szegedy, Scott Reed,
  Cheng-Yang Fu, and Alexander~C. Berg.
\newblock {SSD}: {S}ingle shot multibox detector.
\newblock In {\em Proc. ECCV}, 2016.

\bibitem{neven2018}
Davy Neven, Bert De~Brabandere, Stamatios Georgoulis, Marc Proesmans, and Luc
  Van~Gool.
\newblock Towards end-to-end lane detection: An instance segmentation approach.
\newblock In {\em Intelligent Vehicles Symposium}, 2018.

\bibitem{pan2018}
Xingang Pan, Jianping Shi, Ping Luo, Xiaogang Wang, and Xiaoou Tang.
\newblock Spatial as deep: Spatial {CNN} for traffic scene understanding.
\newblock In {\em Proc. AAAI}, 2018.

\bibitem{park2022}
Wonhui Park, Dongkwon Jin, and Chang-Su Kim.
\newblock Eigencontours: {N}ovel contour descriptors based on low-rank
  approximation.
\newblock In {\em Proc. IEEE CVPR}, 2022.

\bibitem{philion2019}
Jonah Philion.
\newblock Fast{D}raw: {A}ddressing the long tail of lane detection by adapting
  a sequential prediction network.
\newblock In {\em Proc. IEEE CVPR}, 2019.

\bibitem{qin2020}
Zequn Qin, Huanyu Wang, and Xi Li.
\newblock Ultra fast structure-aware deep lane detection.
\newblock In {\em Proc. ECCV}, 2020.

\bibitem{qu2021}
Zhan Qu, Huan Jin, Yang Zhou, Zhen Yang, and Wei Zhang.
\newblock Focus on {L}ocal: Detecting lane marker from bottom up via key point.
\newblock In {\em Proc. IEEE CVPR}, 2021.

\bibitem{redmon2016yolo}
Joseph Redmon, Santosh Divvala, Ross Girshick, and Ali Farhadi.
\newblock You only look once: {U}nified, real-time object detection.
\newblock In {\em Proc. IEEE CVPR}, 2016.

\bibitem{ren2015faster}
Shaoqing Ren, Kaiming He, Ross Girshick, and Jian Sun.
\newblock Faster {R}-{CNN}: {T}owards real-time object detection with region
  proposal networks.
\newblock In {\em Proc. NIPS}, 2015.

\bibitem{savitzky1964}
Abraham Savitzky and Marcel~JE. Golay.
\newblock Smoothing and differentiation of data by simplified least squares
  procedures.
\newblock {\em Analytical Chemistry}, 36(8):1627--1639, 1964.

\bibitem{tabelini2021}
Lucas Tabelini, Rodrigo Berriel, Thiago~M Paixao, Claudine Badue, Alberto~F
  De~Souza, and Thiago Oliveira-Santos.
\newblock Keep your eyes on the lane: Real-time attention-guided lane
  detection.
\newblock In {\em Proc. IEEE CVPR}, 2021.

\bibitem{turk1991face}
Matthew~A. Turk and Alex~P. Pentland.
\newblock Face recognition using eigenfaces.
\newblock In {\em Proc. IEEE CVPR}, 1991.

\bibitem{wang2020poly}
Bingke Wang, Zilei Wang, and Yixin Zhang.
\newblock Polynomial regression network for variable-number lane detection.
\newblock In {\em Proc. ECCV}, 2020.

\bibitem{li2020curvelane}
Hang Xu, Shaoju Wang, Xinyue Cai, Wei Zhang, Xiaodan Liang, and Zhenguo Li.
\newblock Curve{L}ane-{NAS}: {U}nifying lane-sensitive architecture search and
  adaptive point blending.
\newblock In {\em Proc. ECCV}, 2020.

\bibitem{zheng2021}
Tu Zheng, Hao Fang, Yi Zhang, Wenjian Tang, Zheng Yang, Haifeng Liu, and Deng
  Cai.
\newblock {RESA}: Recurrent feature-shift aggregator for lane detection.
\newblock In {\em Proc. AAAI}, 2021.

\bibitem{zhou2010}
Shengyan Zhou, Yanhua Jiang, Junqiang Xi, Jianwei Gong, Guangming Xiong, and
  Huiyan Chen.
\newblock A novel lane detection based on geometrical model and gabor filter.
\newblock In {\em Intelligent Vehicles Symposium}, 2010.

\end{thebibliography}
}

\end{document}